\documentclass{article}

\usepackage[subtle, title=normal, sections=normal, margins=normal]{savetrees}
\usepackage{microtype}
\usepackage{graphicx}
\usepackage{booktabs}
\usepackage{xcolor}
\usepackage{hyperref}
\usepackage{tcolorbox}
\tcbuselibrary{breakable}

\usepackage[accepted]{mlsys2025}
\setcitestyle{numbers,square,comma}

\usepackage{multirow}
\usepackage{array}
\usepackage{tikz}
\usepackage{pgfplots}
\usepackage{listings}
\usepackage{enumitem}
\setlist[enumerate]{nosep, leftmargin=*}
\usepackage{booktabs,tabularx}
\usetikzlibrary{patterns,patterns.meta}
\usepgfplotslibrary{groupplots}

\definecolor{set2cyan}{RGB}{114,182,161}
\definecolor{set2orange}{RGB}{232,149,117}
\definecolor{set2blue}{RGB}{149,163,195}
\definecolor{set2purple}{RGB}{219,149,192}
\definecolor{set2green}{RGB}{162,199,101}
\definecolor{set2yellow}{RGB}{255,217,47}
\definecolor{set2brown}{RGB}{229,196,148}
\definecolor{set2gray}{RGB}{88,88,88}
\definecolor{dark2teal}{RGB}{27,158,119}
\definecolor{dark2orange}{RGB}{217,95,2}
\definecolor{dark2pink}{RGB}{231,41,138}
\definecolor{dark2green}{RGB}{102,166,30}
\definecolor{dark2brown}{RGB}{166,118,29}
\definecolor{set2peach}{RGB}{253,192,134}
\definecolor{tealblue}{RGB}{54,144,192}
\definecolor{softcoral}{RGB}{242,133,108}
\definecolor{amethystpurple}{RGB}{153,102,204}
\colorlet{legendborder}{gray}

\colorlet{cGen}{set2cyan}
\colorlet{cGenAdv}{set2orange}
\colorlet{cRouted}{set2blue}
\colorlet{cRepair}{set2purple}
\colorlet{cEvo}{set2green}
\colorlet{cControl}{set2gray}

\pgfplotsset{
  compat=1.18,
  paperaxis/.style={
    label style={font=\small},
    tick label style={font=\small},
    title style={font=\small, yshift=-3pt},
    ymajorgrids=true, xmajorgrids=false,
    grid style={dashed, lightgray, dash pattern=on 1pt off 1pt},
    xtick align=outside, xtick pos=bottom,
    x tick style={color=black, line width=0.6pt},
    ytick style={draw=none},
    major tick length=1mm,
    axis line style={line width=0.5pt},
    legend style={font=\footnotesize, draw=legendborder, fill=white,
      /tikz/column sep=1.1pt, /tikz/row sep=0pt, inner xsep=3pt, inner ysep=1pt},
    legend cell align={left},
    every axis plot/.append style={line width=1.2pt},
  },
  barplot/axis/.style={
    paperaxis, ybar, bar width=5pt, enlarge x limits=0.14,
    xtick=data, tick style={draw=none},
    xticklabel style={font=\footnotesize},
    legend image code/.code={\draw[##1] (0pt,-0.6ex) rectangle (0.8em,0.6ex);},
  },
  barplot/methods/.style={
    cycle list={
      {bar shift=-2*\pgfplotbarwidth, draw=cGen!60!black, fill=cGen},
      {bar shift=-1*\pgfplotbarwidth, draw=cGenAdv!80!black, fill=cGenAdv!60,
       postaction={pattern={Lines[angle=45,distance=3pt]}, pattern color=cGenAdv!80!black}},
      {bar shift=0*\pgfplotbarwidth, draw=cRouted!80!black, fill=cRouted!45,
       postaction={pattern={Dots[distance=3pt]}, pattern color=cRouted!90!black}},
      {bar shift=+1*\pgfplotbarwidth, draw=cRepair!90!black, fill=cRepair!55,
       postaction={pattern={Hatch[angle=0,distance=3pt]}, pattern color=cRepair!95!black},
       postaction={pattern={Hatch[angle=90,distance=3pt]}, pattern color=cRepair!95!black}},
      {bar shift=+2*\pgfplotbarwidth, draw=cEvo!80!black, fill=cEvo!70,
       postaction={pattern={Hatch[angle=45,distance=3pt]}, pattern color=cEvo!80!black}},
    },
  },
  barplot/two/.style={
    cycle list={
      {bar shift=-0.5*\pgfplotbarwidth, draw=set2cyan!60!black, fill=set2cyan},
      {bar shift=+0.5*\pgfplotbarwidth, draw=amethystpurple!80!black, fill=amethystpurple,
       postaction={pattern={Lines[angle=45,distance=3pt]}, pattern color=amethystpurple!80!black}},
    },
  },
  lGen/.style={color=cGen, solid, line width=1.4pt},
  lGenAdv/.style={color=cGenAdv, dash pattern=on 4pt off 2pt, line width=1.4pt},
  lRouted/.style={color=cRouted, dash pattern=on 2pt off 2pt, line width=1.4pt},
  lRepair/.style={color=cRepair, dash pattern=on 1pt off 1.5pt, line width=1.4pt},
  lEvo/.style={color=cEvo, dash pattern=on 5pt off 1.5pt on 1pt off 1.5pt, line width=1.4pt},
  lFloor/.style={color=set2gray, solid, line width=1pt},
  lControl/.style={color=set2gray, dash pattern=on 4pt off 2pt, line width=1.4pt},
}

\newcommand{\betterV}[3]{%
  \draw[->, line width=1.1pt, set2gray, >=stealth] (rel axis cs:#1,#2) -- (rel axis cs:#1,#3);
  \node[rotate=90, font=\bfseries\scriptsize, text=set2gray, anchor=south, inner sep=1pt]
    at ($(rel axis cs:#1,#2)!0.5!(rel axis cs:#1,#3)$) [xshift=-2pt] {Better};}
    
\newcommand{\betterH}[3]{%
  \draw[->, line width=1.1pt, set2gray, >=stealth] (rel axis cs:#2,#1) -- (rel axis cs:#3,#1);
  \node[font=\bfseries\scriptsize, text=set2gray, anchor=south, inner sep=1pt]
    at ($(rel axis cs:#2,#1)!0.5!(rel axis cs:#3,#1)$) [yshift=2pt] {Better};}

\newcommand{\betterD}[4]{%
  \draw[->, line width=1.1pt, set2gray, >=stealth]
    (rel axis cs:#1,#2) --
    node[midway, sloped, above, font=\bfseries\scriptsize, text=set2gray, inner sep=1pt] {Better}
    (rel axis cs:#3,#4);}
\usetikzlibrary{arrows.meta,positioning,shapes.geometric,fit,calc,backgrounds}

\definecolor{genfill}{HTML}{DCE7F2}   
\definecolor{genline}{HTML}{3D6E9E}
\definecolor{incfill}{HTML}{EDEDED}   
\definecolor{incline}{HTML}{7A7A7A}
\definecolor{gatefill}{HTML}{F6F1E3}  
\definecolor{gateline}{HTML}{A08C5B}
\definecolor{rejline}{HTML}{B03A48}   
\tikzset{
  archnode/.style={rounded corners=1.6pt, draw, line width=.5pt,
                   inner xsep=4pt, inner ysep=3pt, font=\scriptsize,
                   minimum height=13pt, align=center},
  gen/.style   ={archnode, fill=genfill,  draw=genline},
  inc/.style   ={archnode, fill=incfill,  draw=incline},
  gate/.style  ={archnode, fill=gatefill, draw=gateline, inner xsep=3pt},
  route/.style ={archnode, fill=white, draw=genline, trapezium,
                 trapezium left angle=72, trapezium right angle=108,
                 rounded corners=0pt, inner xsep=3pt},
  data/.style  ={archnode, fill=white, draw=black!45, densely dotted,
                 line width=.6pt, font=\scriptsize},
  agent/.style ={archnode, fill=white, draw=genline, line width=.7pt,
                 minimum width=15pt, font=\scriptsize\ttfamily},
  io/.style    ={font=\scriptsize, inner sep=1pt},
  flow/.style  ={-{Stealth[length=3.4pt,width=3pt]}, line width=.5pt},
  faint/.style ={flow, draw=genline!35},
  rej/.style   ={flow, draw=rejline, dashed, dash pattern=on 1.6pt off 1.3pt},
  lbl/.style   ={font=\tiny, inner sep=1.5pt, text=black!65},
  rejlbl/.style={font=\tiny, inner sep=1.5pt, text=rejline},
  panel/.style ={font=\scriptsize\bfseries},
  phase/.style ={font=\tiny\itshape, text=black!55, inner sep=1pt},
}
\newcolumntype{L}[1]{>{\raggedright\arraybackslash}p{#1}}

\usepackage{amsmath}
\usepackage{amssymb}
\usepackage{amsthm}
\usepackage{subcaption}
\usepackage{bm}
\usetikzlibrary{intersections}
\newtheorem{proposition}{Proposition}
\newtheorem{theorem}{Theorem}
\newtheorem{corollary}{Corollary}

\usepackage{xspace}
\newcommand{\name}{\textsc{Tailor}\xspace}

\newcommand{\CVaR}{\mathrm{CVaR}}
\newcommand{\VaR}{\mathrm{VaR}}
\newcommand{\LB}{\mathrm{LB}}
\newcommand{\OPT}{\mathrm{OPT}}
\newcommand{\breq}{\beta_{\mathrm{req}}}   
\newcommand{\bceil}{\beta^{ceil}}           

\makeatletter
\renewcommand{\sectionautorefname}{\S\@gobble}
\renewcommand{\subsectionautorefname}{\S\@gobble}
\renewcommand{\subsubsectionautorefname}{\S\@gobble}
\makeatother

\makeatletter
\renewcommand{\printAffiliationsAndNotice}[1]{%
  \stepcounter{@affiliationcounter}%
  {\let\thefootnote\relax\footnotetext{\hspace*{-\footnotesep}\ifdefined\isaccepted #1\fi%
  \forloop{@affilnum}{1}{\value{@affilnum} < \value{@affiliationcounter}}{%
    \textsuperscript{\arabic{@affilnum}}\ifcsname @affilname\the@affilnum\endcsname%
      \csname @affilname\the@affilnum\endcsname%
    \else
      {\bf AUTHORERR: Missing \textbackslash{}mlsysaffiliation.}%
    \fi
  }.%
  }}%
}
\makeatother

\mlsystitlerunning{Robustness Starts with the Objective: Designing and Certifying Robust LLM-Generated Heuristics}

\begin{document}

\twocolumn[
\mlsystitle{Robustness Starts with the Objective: \\ Designing and Certifying Robust LLM-Generated Heuristics}

\begin{mlsysauthorlist}
\mlsysauthor{Pantea Karimi}{mit}
\mlsysauthor{Roozbeh Bostandoost}{umass}
\mlsysauthor{Dany Rouhana}{msr}
\mlsysauthor{Siva Kesava Reddy Kakarla}{msr}
\mlsysauthor{Pooria Namyar}{msr}
\mlsysauthor{Mohammad Alizadeh}{mit}
\mlsysauthor{Venkat Arun}{utaustin}
\mlsysauthor{Behnaz Arzani}{msr}
\end{mlsysauthorlist}

\mlsysaffiliation{mit}{MIT}
\mlsysaffiliation{umass}{University of Massachusetts Amherst}
\mlsysaffiliation{msr}{Microsoft}
\mlsysaffiliation{utaustin}{The University of Texas at Austin}
\mlsyskeywords{LLM-generated heuristics, certification, conditional value at risk, online algorithms}

\vskip 0.3in

\begin{abstract}
LLMs are increasingly used to design heuristics for computer systems.
However, most synthesis frameworks evaluate candidates by their average
performance, which can mask severe regressions on specific workloads.
Even adding adversarial training inputs doesn’t eliminate these regressions,
and percentile objectives (e.g., optimizing for the 95th percentile) overlook how severe regressions can be beyond the
selected threshold.

We present TAILOR, a framework that guides LLM synthesis using the
Conditional Value at Risk (CVaR) of the cost difference between
each candidate and an existing baseline on the same workload. This paired
CVaR measures the average difference over the
worst fraction of workloads. We show that a negative paired CVaR guarantees improvements in both \emph{average performance} and \emph{tail optimality gaps} relative to the unknown optimum, \emph{without} ever computing that optimum.

We find that TAILOR achieves the best mean--tail trade-off among the synthesis objectives we evaluate across three case studies: Azure GPU batch admission, Azure VM packing, and server placement. Tested across seven diverse optimization benchmarks, TAILOR produces heuristics that are certified with 99\% confidence to improve both average performance and tail optimality gaps over the baseline in six of them. On these six benchmarks, TAILOR also outperforms known robust, human-designed alternatives in the tail while achieving better or comparable average-performance improvement over the baseline.

\end{abstract}
]

\printAffiliationsAndNotice{}

\definecolor{takeawaycolor}{RGB}{54,144,192}  

\newtcolorbox{takeaway}{
  breakable,
  colback=takeawaycolor!7,
  colframe=takeawaycolor!70!black,
  boxrule=0.5pt,
  arc=2pt,
  left=4pt, right=4pt, top=3pt, bottom=3pt,
}

\section{Introduction}
\label{sec:introduction}

Computer systems rely on heuristics to allocate resources,
pack workloads, schedule jobs, and route traffic~\cite{selinger1979access,
megiddo2003arc,graham1969bounds,hong2013swan}. These heuristics aim to
minimize cost or maximize utility when computing an exact optimal solution
($\OPT$) is computationally intractable or requires knowledge of future
inputs. Designing effective heuristics requires substantial
human expertise to exploit problem structure while maintaining reliable
performance across diverse workload regimes.

LLMs have opened new opportunities to automate heuristic design.
Recent frameworks use different search harnesses, including evolutionary
search, orchestrated agents, and iterative refinement, to generate
candidate programs, evaluate them on test inputs, and use the resulting scores
to guide the search~\cite{shinkaevolve2025,karimi2025robusta,
karimi2026engram,glia,romeraparedes2024mathematical,
novikov2025alphaevolve}.

Despite differences in search architecture, most LLM-based heuristic design frameworks optimize for \emph{average} performance and give limited attention to {\em robustness}. FunSearch~\cite{romeraparedes2024mathematical} and EoH~\cite{liu2024eoh} score candidates by their mean performance on a fixed training set, while EoH-S~\cite{liu2026eohs} evolves a portfolio of heuristics over a predefined set of diverse instances. Even recent methods that target robustness do not directly optimize tail or worst-case performance during search. RAISE~\cite{liu2026raise} adds adversarially discovered hard instances to its evaluation set, but still ranks candidates by their \emph{mean} score over that set, arguing that scoring based on the worst lets single outlier instances dominate and destabilize the search~\cite{liu2026raise}. But in production, good average performance can hide severe regressions on specific workloads. For example, MetaOpt~\cite{namyar2024metaopt} identified a traffic pattern on which a production heuristic in Microsoft's WAN served $40\%$ less demand than the optimal solution. 

\begin{figure}[t]
\centering
\def\fsmark{\fontsize{8}{9.6}\selectfont}%
\def\fstag{\fontsize{7}{8.4}\selectfont}%
\begin{tikzpicture}
\begin{axis}[
    scale only axis,
    width=0.86\linewidth,
    height=3.0cm,
    xmin=0, xmax=760,
    ymin=0, ymax=1.28,
    axis lines=left,
    axis line style={black!55, line width=0.8pt},
    ylabel={Probability mass},
    ylabel style={font=\fsmark},
    xtick=\empty, ytick=\empty,
    clip=false,
    legend columns=2,
    legend cell align=left,
    legend style={at={(0.0,1.03)}, anchor=south west, draw=black!25,
                  fill=white, font=\fstag, inner sep=2pt, row sep=-1pt},
]
\addplot[draw=none, fill=softcoral!25, restrict x to domain=496:760, forget plot]
    table[x=x, y=y] {figs/data/intro_real_h.dat} \closedcycle;
\addplot[draw=none, pattern={Lines[angle=45, distance=2.2pt, line width=0.5pt]},
    pattern color=tealblue!80, restrict x to domain=423:760, forget plot]
    table[x=x, y=y] {figs/data/intro_real_hprime.dat} \closedcycle;

\addplot[softcoral, solid, line width=1.6pt, smooth, name path=hcurve]
    table[x=x, y=y] {figs/data/intro_real_h.dat};
\addlegendentry{$H-\OPT$ (baseline)}
\addplot[tealblue, dashed, line width=1.6pt, smooth, name path=hpcurve]
    table[x=x, y=y] {figs/data/intro_real_hprime.dat};
\addlegendentry{$H'-\OPT$ (candidate)}

\draw[-{Latex[length=1.8mm]}, black!45, line width=0.8pt]
     (axis cs:130,-0.10) -- (axis cs:20,-0.10);
\node[font=\fstag, text=black!55, anchor=west] at (axis cs:138,-0.10) {better};

\node[font=\fsmark, text=black!70, anchor=north] at (axis cs:290,-0.20) {Optimality gap};

\path[name path=vhp] (axis cs:423,0) -- (axis cs:423,1.28);
\path[name path=vh]  (axis cs:496,0) -- (axis cs:496,1.28);
\path[name intersections={of=hpcurve and vhp, by=hpTop}];
\path[name intersections={of=hcurve and vh,  by=hTop}];
\draw[tealblue!90!black,  densely dotted, line width=1.1pt] (axis cs:423,0) -- (hpTop);
\draw[softcoral!90!black, densely dotted, line width=1.1pt] (axis cs:496,0) -- (hTop);
\node[font=\fstag, text=softcoral!90!black, anchor=north, align=center] (varlab)
     at (axis cs:590,-0.10) {worst-5\% cutoff};
\draw[-{Latex[length=1.4mm]}, softcoral!85!black, line width=0.6pt]
     (varlab.north) -- (axis cs:496,0.0);

\draw[tealblue!90!black,  line width=1.3pt] (axis cs:462,0) -- (axis cs:462,0.52);
\draw[softcoral!90!black, line width=1.3pt] (axis cs:537,0) -- (axis cs:537,0.52);
\node[font=\fstag, text=softcoral!90!black, anchor=south, align=center] (cvarlab)
     at (axis cs:640,0.62) {average of \\ worst 5\%};
\draw[-{Latex[length=1.4mm]}, softcoral!85!black, line width=0.6pt]
     (cvarlab.south) -- (axis cs:539,0.50);

\draw[-{Latex[length=1.6mm]}, black!70, line width=0.9pt]
     (axis cs:537,0.84) -- (axis cs:462,0.84);
\node[font=\fstag, text=black!75, anchor=south, align=center] at (axis cs:500,0.9)
     {$H'$ lowers the worst-5\% \\ average by 14\%};
\end{axis}
\end{tikzpicture}
\caption{\textbf{Goal: improve tail optimality gaps while
preserving average performance.}
Distributions of optimality gaps $H-\OPT$ and
$H'-\OPT$ for GPU batch admission (\autoref{sec:motivation}),
where $H$ and $H'$ denote the baseline and candidate
performance, expressed as costs to minimize.
The shaded regions are the worst 5\% of each
distribution, starting at its 95th-percentile cutoff.
$H'$ reduces the tail optimality gap
by 14\% while preserving average performance.}
\vspace{-5 pt}
\label{fig:motivation-tail}
\label{fig:motivation}
\end{figure}
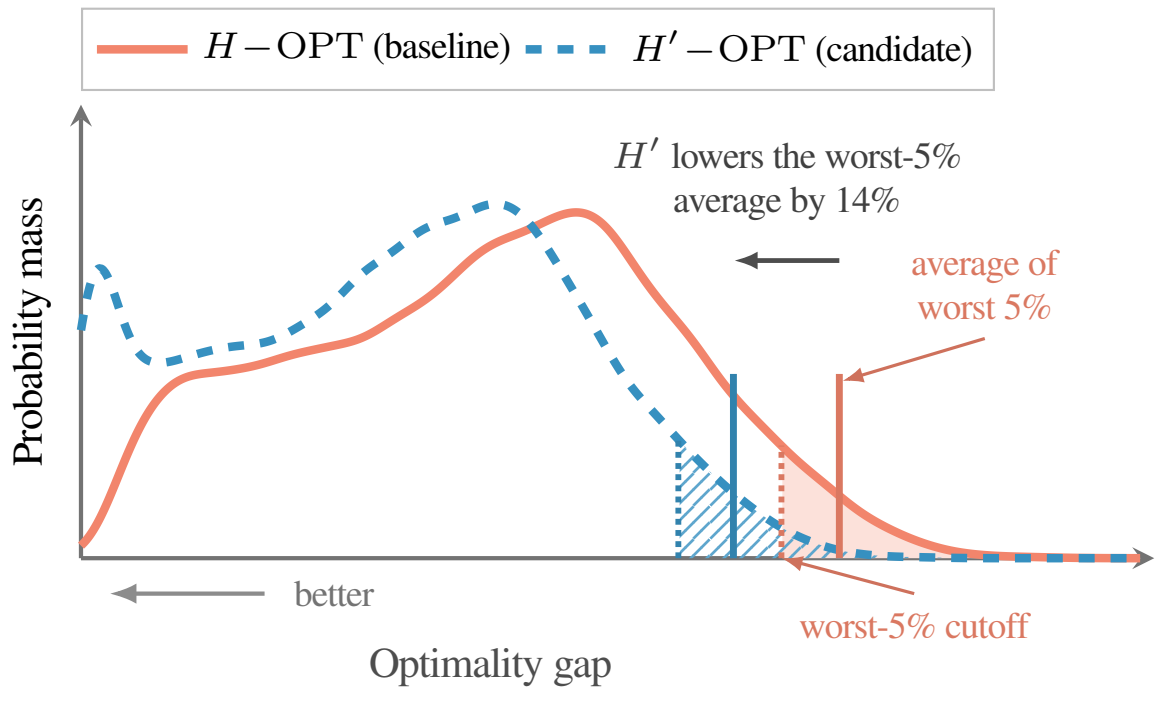

In this paper, we seek to automate the design of
\emph{robust heuristic improvement}: improving tail
performance without sacrificing average performance.
Let $I \sim \mathcal{D}$ denote a workload drawn from
a target distribution $\mathcal{D}$, $H$ an existing
baseline heuristic, and $H'$ a replacement generated by
an LLM. We write $H(I)$ and $H'(I)$ for their performance
on the same workload, expressed as costs to minimize
(maximization problems are handled by negating rewards).
For example, in GPU Batch Admission
(\autoref{sec:motivation}), requests arrive
sequentially and compete for limited batch capacity.
Here, $I$ is a stream of requests, $H$ could be
first-come, first-served (FCFS), and $H'$ a candidate LLM-generated policy.
Their performance, $H(I)$ and $H'(I)$, is the negative
of the total value admitted on that stream.
$\OPT(I)$ denotes the best possible cost had all
future requests been known in advance.
The \emph{optimality gap}, $H(I)-\OPT(I)$, measures
how much value the baseline leaves on the table.
A heuristic may perform well on average while
suffering large optimality gaps on a fraction
of workloads.
Our goal is to find a replacement $H'$ via LLMs that reduces
the tail of these optimality gaps while preserving
or improving average performance over $\mathcal{D}$
(\autoref{fig:motivation-tail}).

To control tail regressions, we would ideally optimize
the tail of the candidate's optimality gaps directly
(\autoref{fig:motivation-tail}).
However, the challenge is that computing $\OPT$ across large-scale workloads is impractical.
Existing tools such as MetaOpt~\cite{namyar2024metaopt}
and MetaEase~\cite{karimi2026metaease} identify worst-case
failures relative to $\OPT$, but require formal models
or expensive $\OPT$ evaluations, taking minutes to hours
even on small instances (\autoref{app:opt}). Other tools have similar problems~\cite{minaQueue, venkat,ccac,hotcocoa,buffy,xplain,sched-perf-verify}.
This makes them impractical for iterative LLM synthesis,
which must evaluate thousands of candidate heuristics.
A simpler alternative, optimizing the tail of $H'$
directly (e.g., its 95th-percentile cost), is also
insufficient (\autoref{sec:motivation}).
A high cost may reflect an inherently difficult workload
rather than poor heuristic performance, as even an
optimal solution may incur a high cost.
We therefore need a scalable objective that measures
tail performance relative to the baseline without
computing $\OPT$.

The second challenge is the tension between average
performance and tail robustness, which arises in many
problems~\cite{zhou2026aichilles}.
In GPU Batch Admission, optimizing average
performance leads LLMs to discover heuristics that achieve
large average improvements while incurring severe
regressions on some workloads.
We need an objective that accounts for how often regressions occur and how severe, without
sacrificing average performance.

\noindent\textbf{Our approach: Optimizing the tail of paired differences.}
We introduce \name, a framework for synthesizing robust
heuristic improvements with LLMs.
Rather than computing $\OPT$, we evaluate the candidate
and baseline on the same workload and compute their
\emph{paired difference}, $\Delta(I) = H'(I) - H(I)$.
This directly measures the impact of replacing the baseline
and requires only running the two heuristics, making it
practical feedback for iterative LLM search.

Even with paired differences, simple tail metrics provide
poor feedback for LLM search.
Optimizing a fixed percentile (e.g., the 95th percentile
of $\Delta$) ignores how severe regressions get beyond the cutoff,
while using the single worst-case difference lets isolated
outliers dominate heuristic selection (\autoref{sec:maxgap}).
We instead minimize the Conditional Value-at-Risk (CVaR)
of paired differences
~\cite{rockafellar2000optimization,rockafellar2002conditional}.
At $\beta=0.95$, $\CVaR_\beta(\Delta)$ averages the worst
5\% of candidate--baseline differences, 
capturing both how often regressions occur and how severe they
are. 

\noindent\textbf{Certifying tail improvements without OPT.}
We show a negative population
$\CVaR_\beta(\Delta)$ guarantees that replacing the
baseline $H$ with $H'$ reduces the average optimality
gap in the worst $1-\beta$ tail of the workload
distribution.
This guarantees an improvement relative to $\OPT$
without ever computing $\OPT$, using only the observable
performance differences between the two heuristics
(\autoref{sec:problem-formulation}).

\noindent\textbf{Guaranteeing average improvement.}
CVaR has a second useful property: the average of the
worst performance differences is always at least as
large as their overall average.
Therefore, if even the worst-tail average difference
is negative, the overall average difference must also
be negative.
\emph{A negative paired CVaR therefore guarantees
average improvement, without explicitly optimizing
for the mean or imposing a separate constraint.}

This guarantee holds for the full distribution, but in practice we can only evaluate a candidate on a finite sample. To account for sampling uncertainty,
\name\ evaluates the final heuristic on independent
validation workloads and computes a statistical
upper confidence bound on its paired CVaR~\cite{massart1990tight}.
If this bound is negative, we certify with 99\%
confidence that the heuristic improves both tail
optimality gaps and average performance over the
target distribution. We also introduce a simple check to detect when
it is unlikely that we would be able to certify
an improvement before synthesis. This enables us
to avoid the expensive LLM-based synthesis when the
requested certification target is unattainable
(\autoref{sec:problem-formulation}).

We evaluate \name\ across three case studies on realistic
workloads: Azure GPU Batch Admission~\cite{stojkovic2025dynamollm,azure2024llmtrace},
Azure VM Packing~\cite{azurepackingtrace2020,hadary2020protean},
and Server Placement~\cite{qiu2001placement}.
Holding the LLM and synthesis architecture fixed, we
compare different synthesis objectives, measuring each
heuristic's average improvement and its performance
relative to the baseline on the worst 5\% of paired
differences.

\textbf{\name\ improves tail performance over mean-optimized heuristics.}
On GPU Batch Admission, mean-optimized heuristics underperform the FCFS baseline on $7.5\%$ of streams, with an average regression of $9.2\%$ over the worst $5\%$ of streams.
Adding adversarial inputs does not eliminate regressions: RAISE~\cite{liu2026raise} still regresses the baseline on $7.4\%$ of streams. \name\ regresses on fewer than $0.01\%$ of streams (\autoref{sec:motivation}).
On Server Placement, \name turns an average regression over the worst $5\%$ of inputs into an improvement.
On VM Packing, it delivers a $25\%$ larger average improvement over the worst $5\%$ of inputs than mean-optimized heuristics (\autoref{sec:more-cases}).

\textbf{\name\ improves tail performance more than percentile-based optimization.}
Across all three case studies, \name\ delivers $46$--$62\%$ larger average improvements over the baseline on the worst $5\%$ of inputs than heuristics synthesized by optimizing the $95$th percentile of paired differences (\autoref{sec:motivation},\autoref{sec:more-cases}).

\textbf{\name\ certifies its tail improvements.}
Across the three case studies and four additional synthesis
benchmarks, \name\ certifies improvements in tail
optimality gaps in 24 of 35 independent synthesis
runs at 99\% confidence, with successful certification
on six of the seven benchmarks
(\autoref{sec:eval-generalization}).

\textbf{\name\ outperforms robust human-designed heuristics in the tail.}
Across six benchmarks with known robust human-designed heuristics, the best \name-generated heuristics achieve a better worst-5\% average relative to the same baseline.
On Traffic Engineering and VM Packing, human-designed
heuristics regress on average in their worst 5\% of
outcomes, while \name-generated heuristics still
improve over the baseline (\autoref{sec:eval-best}).

\textbf{\name\ also improves average performance.}
Across all cases, \name achieves better average performance over the baseline heuristic.
Also on VM Packing and Server Placement, \name\ beats mean-optimized heuristics and achieves $14\%$ and $66\%$ larger average improvements.

\definecolor{azpurple}{RGB}{117,112,179}
\tikzset{
  muted/.style={line width=1.5pt},
  objMean/.style={muted, dashed,          color=dark2orange!55},
  objMeanH/.style={muted, dash dot,       color=dark2brown!55},
  objPctH/.style={muted, dotted,          color=dark2pink!55},
  objPct/.style={muted, densely dashed,   color=dark2teal!55},
  objPtail/.style={muted, densely dotted, color=tealblue!55},
  objAttack/.style={muted, densely dashdotdotted, color=red!70!black!55},
  objRaise/.style={muted, dash dot dot,   color=azpurple!55},
  objZhou/.style={muted, loosely dashed,  color=set2gray},
  objOurs/.style={solid, line width=2.2pt, color=dark2green},
}

\begin{figure*}[t]
    \centering
    \definecolor{azpurple}{RGB}{117,112,179}
\definecolor{pctblue}{RGB}{31,120,180}   
\definecolor{pctsky}{RGB}{86,180,233}    
\definecolor{pctgold}{RGB}{230,171,2} 

\newcommand{\drcdfguides}{%
  \draw[black!60, line width=0.4pt] (axis cs:0,\pgfkeysvalueof{/pgfplots/ymin}) -- (axis cs:0,\pgfkeysvalueof{/pgfplots/ymax});
  \draw[black!75, line width=0.6pt, dash pattern=on 3pt off 2pt]
    (axis cs:\pgfkeysvalueof{/pgfplots/xmin},2.94444) -- (axis cs:\pgfkeysvalueof{/pgfplots/xmax},2.94444);
}

\begin{tikzpicture}[
  muted/.style={line width=1.5pt},
  objMean/.style={muted, dashed,          color=dark2orange!55},
  objMeanH/.style={muted, dash dot,       color=dark2brown!55},
  objPctH/.style={muted, dotted,          color=dark2pink!55},
  objPct/.style={muted, densely dashed,   color=pctgold!75},
  objPtail/.style={muted, densely dotted, color=pctsky},
  objAttack/.style={muted, densely dashdotdotted, color=red!70!black!55},
  objRaise/.style={muted, dash dot dot,   color=azpurple!55},
  objZhou/.style={muted, loosely dashed,  color=set2gray},
  objOurs/.style={solid, line width=2.2pt, color=dark2green}]
\begin{groupplot}[
  group style={group size=2 by 1, horizontal sep=8pt,
               yticklabels at=edge left, ylabels at=edge left},
  paperaxis, scale only axis, width=0.43\linewidth, height=0.26\linewidth,
  ymin=-9.2102, ymax=9.2102,
  ytick={-6.90675,-4.59512,-2.19722,0.00000,2.94444,4.59512,6.90675},
  yticklabels={0.001,0.01,0.1,0.5,0.95,0.99,0.999},
  ylabel={CDF}]

\nextgroupplot[xmin=-106, xmax=31, xtick={-100,-80,-60,-40,-20,0,20},
  legend to name=drcdflegend, legend columns=3, legend cell align=left,
  legend style={font=\scriptsize, draw=black!25, inner sep=1.5pt,
    nodes={inner sep=1pt}, /tikz/column sep=6pt}]
\drcdfguides
\node[font=\scriptsize\bfseries, anchor=north west] at (rel axis cs:0.01,0.97) {(a) Average objectives};
\node[font=\scriptsize, anchor=north west] at (axis cs:1.1,-3.89182) {regressions $\rightarrow$};
\node[font=\scriptsize, anchor=north east] at (axis cs:-1.1,-3.89182) {$\leftarrow$ improvements over FCFS};
\addlegendimage{objMeanH}  \addlegendentry{Mean $H'$}
\addlegendimage{objMean}   \addlegendentry{Mean $\Delta$}
\addlegendimage{objAttack} \addlegendentry{Mean $\Delta$ + attacker}
\addlegendimage{objPctH}   \addlegendentry{95th pct.\ of $H'$}
\addlegendimage{objPct}    \addlegendentry{95th pct.\ of $\Delta$}
\addlegendimage{objPtail}  \addlegendentry{99th pct.\ of $\Delta$}
\addlegendimage{objRaise}  \addlegendentry{RAISE~\cite{liu2026raise}}
\addlegendimage{objZhou}   \addlegendentry{Zhou et al.~\cite{zhou2008budget}}
\addlegendimage{objOurs}   \addlegendentry{Mean worst-5\% of $\Delta$ ($\CVaR(\Delta)$, ours)}
\addplot[objMeanH,  forget plot] table[x=x, y=y] {figs/data/azm_drcdf_avg_pprime.dat};
\addplot[objMean,   forget plot] table[x=x, y=y] {figs/data/azm_drcdf_mean.dat};
\addplot[objAttack, forget plot] table[x=x, y=y] {figs/data/azm_drcdf_mean_attackgap.dat};
\addplot[objRaise,  forget plot] table[x=x, y=y] {figs/data/azm_drcdf_raise.dat};
\addplot[objOurs,   forget plot] table[x=x, y=y] {figs/data/azm_drcdf_cvar.dat};
\betterH{0.85}{0.7}{0.6}

\nextgroupplot[xmin=-35, xmax=16, xtick={-30,-20,-10,0,10}]
\drcdfguides
\node[font=\scriptsize\bfseries, anchor=north west] at (rel axis cs:0.01,0.97) {(b) Tail objectives};
\node[font=\scriptsize, anchor=south west] at (axis cs:-34,2.94444) {worst 5\% of streams $\uparrow$};
\addplot[objPctH,  forget plot] table[x=x, y=y] {figs/data/azm_drcdf_p95_pprime.dat};
\addplot[objPct,   forget plot] table[x=x, y=y] {figs/data/azm_drcdf_p95_delta.dat};
\addplot[objPtail, forget plot] table[x=x, y=y] {figs/data/azm_drcdf_p99_delta.dat};
\addplot[objZhou,  forget plot] table[x=x, y=y] {figs/data/azm_drcdf_zhou.dat};
\addplot[objOurs,  forget plot] table[x=x, y=y] {figs/data/azm_drcdf_cvar.dat};
\end{groupplot}

\node[anchor=south] at ($(group c1r1.north)!0.5!(group c2r1.north)+(0,3pt)$)
  {\pgfplotslegendfromname{drcdflegend}};
\node[anchor=north, font=\small] at ($(group c1r1.outer south)!0.5!(group c2r1.outer south)$)
  {$\Delta = H' - H$ (\% of FCFS average value per stream)};
\end{tikzpicture}
\caption{\textbf{Mean objectives hide tail regressions, and percentile
objectives control only the tail cutoff; paired CVaR controls both.}
CDF of paired differences $\Delta=H'-H$ on held-out Azure streams,
normalized by FCFS average value per stream. Negative values are
improvements over FCFS and positive values are regressions; curves
further left are better. The horizontal line at 0.95 marks the start
of the worst 5\% of streams.
\textbf{(a)}~Mean-based objectives, with or without adversarial
inputs, achieve large improvements on most streams but leave a long
regression tail.
\textbf{(b)}~Percentile objectives control only their tail cutoff,
not the severity of regressions beyond it. Paired CVaR accounts
for the entire tail, achieving better worst-5\% performance
while retaining substantial average improvements.}
\label{fig:motivation-delta-cdf}
\end{figure*}
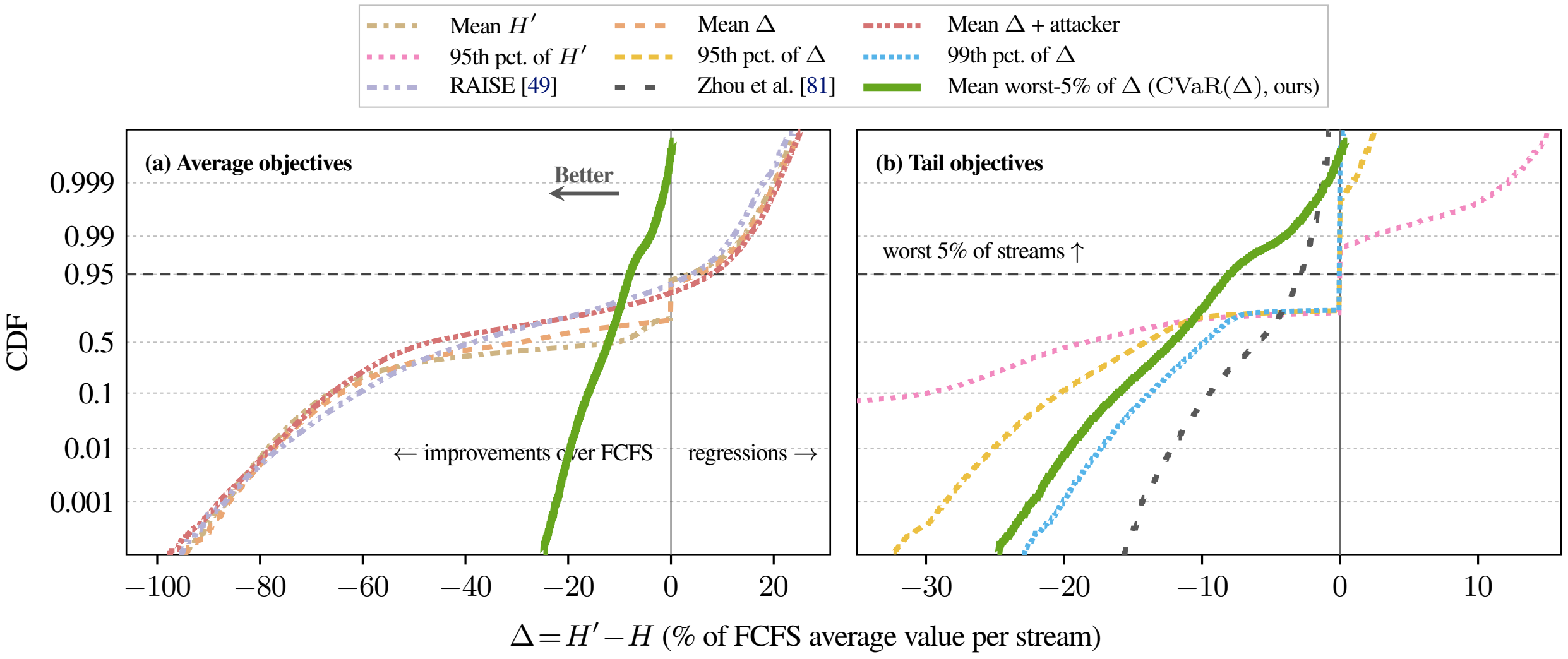

\section{Motivation: What Should We Optimize for Robustness?}
\label{sec:motivation}

\textit{What should an LLM optimize when designing a replacement for an existing
heuristic, when we aim to be more robust-in-tail while maintaining the average performance?} 

We discuss this through a running example: \emph{Azure GPU Batch Admission},
using streams of requests from Azure
LLM-inference traces~\cite{stojkovic2025dynamollm,azure2024llmtrace}.
We model admission as an online knapsack problem:
each request has a value proportional to its token count and a weight given by the batch capacity it consumes, and the batch has a fixed capacity.
Requests arrive sequentially,
and the heuristic must accept or reject each request without knowing
how many more will arrive. The goal is to maximize the total
value of admitted requests, which we use as a proxy for GPU
utilization.

In this example, baseline $H$ is first-come, first-served (FCFS). A generated heuristic $H'$ can do better by passing over low-value
requests in anticipation of more valuable arrivals. The risk is that the
stream might end before those arrivals appear, leaving capacity unused. We express the performance of $H$ and $H'$ as a minimization. 

Ideally, we would evaluate each new heuristic $H'$ against
the optimum. However, computing $\OPT$ is expensive even
on modest-sized instances~\cite{namyar2024metaopt}.
OPT-based synthesis feedback is therefore computationally
prohibitive. We therefore compare a candidate to the baseline. On the same stream $I$, we write
$\Delta(I)=H'(I)-H(I)$; so $\Delta<0$ is an improvement and $\Delta>0$ is a regression.

We examine which synthesis objectives produce heuristics that
improve on the baseline without severe regressions. Starting
with mean optimization, we investigate whether harder training
inputs or increasingly tail-aware objectives lead to more robust
replacements.

We use Engram~\cite{karimi2026engram} with Claude Opus~5 as
the backend model to synthesize heuristics. We train on the first
two calendar days of Azure traces and evaluate on five later,
held-out days, with no overlapping requests. For each objective,
we run five independent syntheses, holding the model, harness,
workloads, and search budget fixed. We additionally evaluate
RAISE~\cite{liu2026raise}, a recent approach that incorporates
adversarial instance search.
\autoref{tab:motivation-ladder} summarizes them,
while \autoref{fig:motivation-delta-cdf} shows their paired-difference
distributions.

\begin{table}[t]
    \centering
    \caption{\textbf{Paired-CVaR balances mean cost improvement and tail robustness.}
    Comparison of synthesis objectives on Azure GPU Batch Admission,
evaluated on our target workload distribution of real Azure
request streams. Results are mean $\pm$ standard deviation over
five independent synthesis runs. Mean cost improvement ($\mathbb{E}[H - H']$) and tail differences are normalized by the magnitude of
    the baseline's average cost. $\Delta = H' - H$.}
    \label{tab:motivation-ladder}
    \scriptsize
    \setlength{\tabcolsep}{4pt}
    \resizebox{\columnwidth}{!}{%
    \begin{tabular}{@{}lrrr@{}}
    \toprule
    Synthesis objective
        & \shortstack{Mean cost\\improvement (\%) $\uparrow$}
        & \shortstack{Regressions\\(\%) $\downarrow$}
        & \shortstack{Worst-5\%\\mean (\%) $\downarrow$} \\
    \midrule
    Mean $H'$
        & $25.7\pm20.8$ & $6.4\pm5.8$ & $+7.4\pm7.9$ \\
    Mean $\Delta$
        & $29.8\pm17.3$ & $7.5\pm5.0$ & $+9.2\pm6.0$ \\
    Mean $\Delta$ + attacker
        & $39.0\pm1.7$ & $10.4\pm1.5$ & $+12.2\pm2.2$ \\
    RAISE~\cite{liu2026raise}
        & $35.5\pm3.4$ & $7.4\pm2.8$ & $+8.6\pm3.1$ \\
    95th pct. of $H'$
        & $17.2\pm10.6$ & $1.6\pm1.9$ & $-0.6\pm4.2$ \\
    95th pct. of $\Delta$
        & $12.3\pm6.9$ & $0.2\pm0.2$ & $-4.1\pm2.3$ \\
    99th pct. of $\Delta$
        & $8.7\pm4.9$ & $<0.01$ & $-4.6\pm2.6$ \\
    \midrule
    Zhou et al.~\cite{zhou2008budget} (human)
        & 6.2 & 0.0 & $-2.4$ \\
    \midrule
    Mean worst 5\% of $\Delta$ (ours)
        & $\mathbf{12.3\pm0.8}$ & $\mathbf{<0.01}$
        & $\mathbf{-6.0\pm0.3}$ \\
    \bottomrule
    \end{tabular}%
    }
\end{table}

\noindent\textbf{Start with the average.}
Most LLM-based heuristic design frameworks optimize average
performance~\cite{romeraparedes2024mathematical,ye2024reevo,
liu2024eoh,novikov2025alphaevolve,cheng2025barbarians}.
We compare minimizing the candidate's mean cost, $\mathbb{E}[H']$,
and its mean difference from the baseline, $\mathbb{E}[\Delta]$,
which differ by a constant for a fixed workload distribution.
Both achieve substantial average improvements, yet regress against FCFS
on 6--8\% of Azure streams (\autoref{fig:motivation-delta-cdf},
\autoref{tab:motivation-ladder}).

The striking observation is \emph{where} the regressions occur. These regressions are concentrated on short streams
(\autoref{fig:motivation-regressions-by-length}).
The reason lies in the strategy the LLM discovers:
the mean-optimized heuristic (\autoref{fig:motivation-code}, left)
aggressively rejects low-value requests in anticipation of more valuable
future arrivals. 
On long streams, the heuristic can reject many low-value requests and still fill the batch. On short streams, however, each rejection risks leaving capacity unused, because few requests remain to replace it. Since the heuristic does not know when
the stream will end, improvements on longer streams can outweigh
severe regressions on shorter ones, making this risky strategy
attractive under mean optimization.

\begin{figure}[t]
    \centering
\begin{tikzpicture}
\begin{axis}[paperaxis, width=0.47\textwidth, height=0.230\textwidth,
  every axis plot/.append style={line width=1.3pt, no markers},
  xmin=220, xmax=640, xtick={220,300,400,500,640},
  xlabel={Stream length $n$}, ylabel={Regressions (\%)},
  ymin=0, ymax=2.0043, ytick={0.00000,0.30103,1.04139,2.00432}, yticklabels={0,1,10,100},
  legend columns=2, legend cell align=left,
  legend style={font=\scriptsize, at={(0.5,1.03)}, anchor=south, /tikz/column sep=4pt}]

\addlegendimage{objMean}  \addlegendentry{Mean $\Delta$}
\addlegendimage{objMeanH} \addlegendentry{Mean $H'$}
\addlegendimage{objAttack} \addlegendentry{Mean $\Delta$ + attacker}
\addlegendimage{objPctH}  \addlegendentry{95th pct.\ of $H'$}
\addlegendimage{objPct}   \addlegendentry{95th pct.\ of $\Delta$}
\addlegendimage{objPtail} \addlegendentry{99th pct.\ of $\Delta$}
\addlegendimage{objRaise} \addlegendentry{RAISE}
\addlegendimage{objZhou}  \addlegendentry{Zhou et al.}
\addlegendimage{objOurs}  \addlegendentry{Mean worst-5\% of $\Delta$ ($\CVaR(\Delta)$, ours)}

\addplot[objMean,  forget plot] table[x=n, y=ylog] {figs/data/azm_mech_mean.dat};
\addplot[objMeanH, forget plot] table[x=n, y=ylog] {figs/data/azm_mech_avg_pprime.dat};
\addplot[objPctH,  forget plot] table[x=n, y=ylog] {figs/data/azm_mech_p95_pprime.dat};
\addplot[objPct,   forget plot] table[x=n, y=ylog] {figs/data/azm_mech_p95_delta.dat};
\addplot[objPtail, forget plot] table[x=n, y=ylog] {figs/data/azm_mech_p99_delta.dat};
\addplot[objAttack, forget plot] table[x=n, y=ylog] {figs/data/azm_mech_mean_attackgap.dat};
\addplot[objRaise, forget plot] table[x=n, y=ylog] {figs/data/azm_mech_raise.dat};
\addplot[objZhou,  forget plot] table[x=n, y=ylog] {figs/data/azm_mech_zhou.dat};
\addplot[objOurs,  forget plot] table[x=n, y=ylog] {figs/data/azm_mech_cvar.dat}; 
\betterV{0.9}{0.62}{0.18}
\end{axis}
\end{tikzpicture}
    \caption{\textbf{Mean-optimized heuristics fail predominantly on short
    streams.} Regression rate against FCFS as a function of stream length.
    The mean-oriented curves rise sharply on the shortest streams, while
    the paired-tail objectives have much lower regression rates. Both
    classes of heuristics improve on nearly all longer streams.}
    \label{fig:motivation-regressions-by-length}
    \label{fig:motivation-mechanism}
    \vspace{-10 pt}
\end{figure}
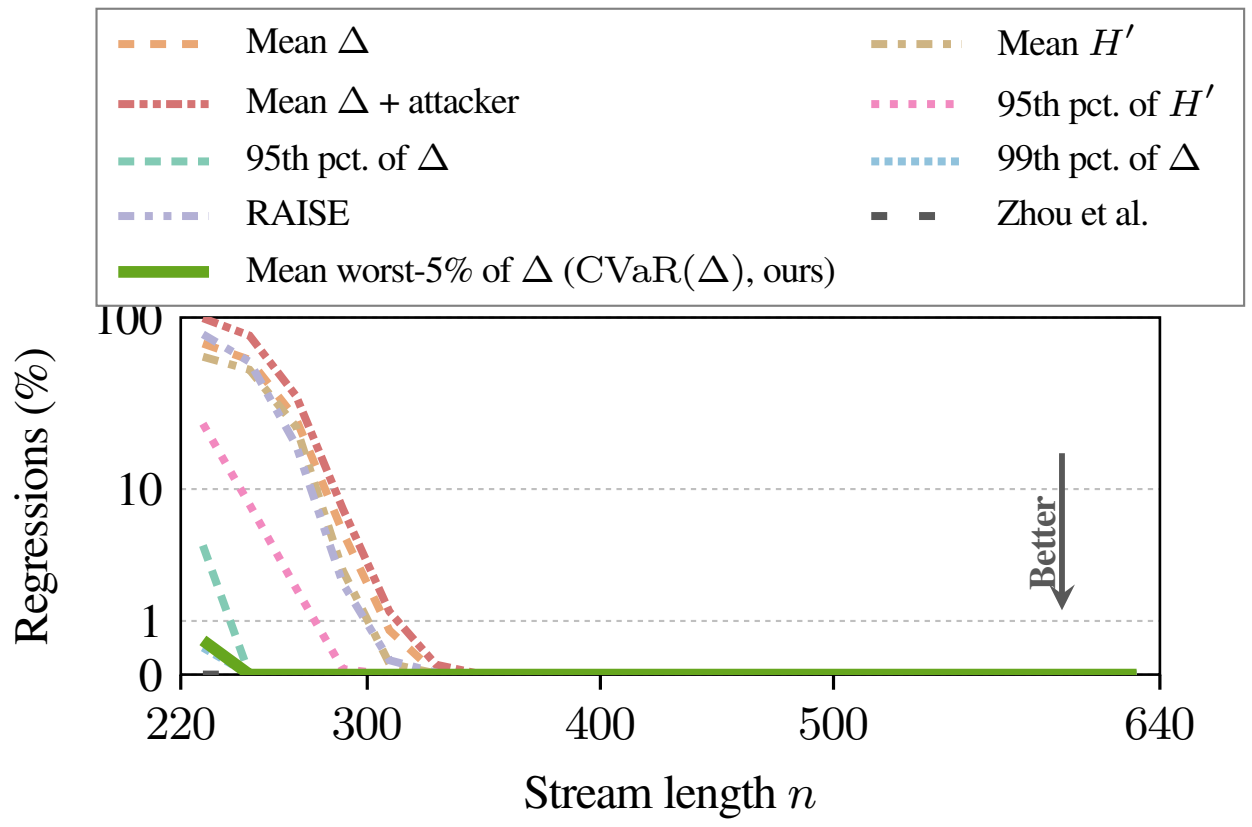

\begin{figure*}[t]
    \centering
    \begin{minipage}[t]{0.485\textwidth}
        \textbf{Mean objective: forecast remaining arrivals}
        \begin{lstlisting}
remaining = (640 - index) / 2   # expected arrivals still to come
if slots_left >= remaining:     # room for everything: accept
    return True
slack = slots_left / remaining  # fewer slots -> higher bar
if slack >= 0.618: return value >= 50_000
# ... intermediate tiers ...
return value >= 400_000
        \end{lstlisting}
        {\footnotesize \emph{Wait:} keep slots available when many
        future requests are expected.}
    \end{minipage}\hfill
    \begin{minipage}[t]{0.485\textwidth}
        \textbf{Paired-CVaR objective: limit early rejection}
        \begin{lstlisting}
fits_left = capacity_left // weight   # copies of this item that still fit
rejected  = index - accepted
if value <= weight and rejected < 22:          # low-value item, rejection budget left: skip
    return False
if fits_left <= 1 and value <= 250_000:        # last slot: save it for a high-value item
    return False
# ... more rules ...
return True
        \end{lstlisting}
        {\footnotesize \emph{Limit the bet for future requests:} stop rejecting the
        cheapest requests indefinitely.}
    \end{minipage}
    \caption{\textbf{The objective changes how the LLM handles uncertainty
    about future arrivals.} Excerpts from generated heuristics per objective. The mean-optimized heuristic derives its threshold
    from a forecast of remaining arrivals (left). The paired-CVaR heuristic
    tracks how many requests it has rejected and uses remaining capacity
    to govern later rejections (right).}
    \label{fig:motivation-code}
\end{figure*}

\noindent\textbf{What if we search for harder inputs?}
Recent work suggests that mean optimization may fail in the
tail because training workloads do not sufficiently expose
the agents to difficult cases~\cite{karimi2025robusta,zhou2026aichilles}.
We test whether exposing the agent to harder workloads
can prevent these regressions without changing the
mean-based synthesis objective. We consider two approaches: RAISE~\cite{liu2026raise}, which generates adversarial
workloads by perturbing request values while preserving
stream lengths, and Mean $\Delta$ + attacker, which searches
for difficult, unmodified Azure streams of varying lengths.

Neither eliminates regressions: RAISE and Mean $\Delta$ +
attacker regress on 7.37\% and 10.40\% of held-out streams,
respectively (\autoref{tab:motivation-ladder}).
Finding difficult inputs does
not determine how to weigh them
against improvements elsewhere. 

\noindent\textbf{What if we optimize the tail percentiles?}
We first minimize the candidate's 95th-percentile cost.
Although this reduces regressions
(\autoref{tab:motivation-ladder}), a high cost of $H'$
may simply reflect a difficult workload rather than
poor performance relative to the baseline.
Optimizing the 95th percentile of $\Delta$ further
reduces regressions, but only controls the tail cutoff and ignores the regressions beyond it.
Moving to the 99th percentile targets rarer regressions,
but reduces average cost improvement from 12.3\% to 8.7\%.
Neither percentile captures the severity of regressions
throughout the tail.

\noindent\textbf{Our approach: Look at the whole tail, not only its percentile.}
These results motivate an objective that accounts for the severity of tail regressions while preserving average performance.
Unlike a percentile, Conditional Value-at-Risk (CVaR) captures the magnitude of losses throughout the tail.
We therefore minimize $\mathrm{CVaR}_{0.95}(\Delta)$, the average of the worst 5\% of paired differences
\cite{rockafellar2000optimization,bogle2019teavar}.
If this average is negative, the candidate improves on the baseline even on its worst 5\% of paired outcomes, which implies improvement on average across the full workload distribution.
Thus, a negative population CVaR provides a sufficient condition for average improvement while avoiding severe tail regressions.

This changes the strategy the LLM discovers. Rather than
continually waiting for better arrivals, CVaR-generated
policies \emph{limit how much they wait}. The heuristic in
\autoref{fig:motivation-code} (right) limits early rejections
and adapts to remaining capacity, substantially reducing
short-stream regressions
(\autoref{fig:motivation-regressions-by-length}).
Interestingly, it independently discovers a mechanism
similar to the human-designed heuristic of Zhou et al.~
\cite{zhou2008budget}, which also limits rejections.
While the human heuristic achieves a 6.2\% average cost improvement
with no observed regressions, CVaR-generated policies
achieve a 12.3\% cost improvement with rare regressions
(\autoref{tab:motivation-ladder}).
Their measured tail improvements also extend beyond
the 5\% level used during synthesis
(\autoref{fig:motivation-tail-sweep}).

These results suggest that robust heuristic design requires
an objective that measures both baseline-relative performance
and the severity of tail losses. However, observed improvements
alone do not establish robustness on unseen workloads.
This motivates \name, which uses the same paired-CVaR objective
not only to guide heuristic synthesis, but after synthesis, to statistically certify improvements in tail optimality gaps, without computing $\OPT$.
We formalize this connection in the next section.

\section{\name: Tail-Aware LLM Optimization for Robustness} 
\label{sec:problem-formulation}

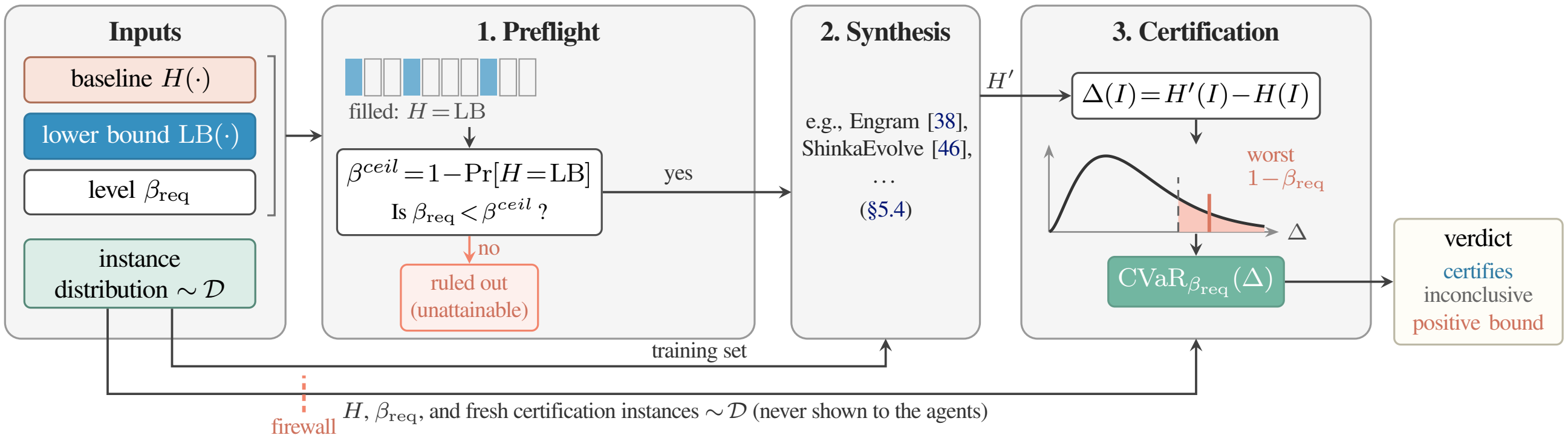
\begin{figure*}[tp]
\centering
\resizebox{\linewidth}{!}{\def\fsbox{\fontsize{8}{9.6}\selectfont}%
\def\fslab{\fontsize{7}{8.4}\selectfont}%
\def\fshead{\fontsize{8.5}{10}\selectfont\bfseries}%
\begin{tikzpicture}[
  font=\fsbox,
  panel/.style={draw=black!40, fill=black!4, rounded corners=5pt, line width=0.6pt},
  head/.style={font=\fshead, text=black!85, anchor=north, inner sep=0pt},
  bx/.style={draw=#1!75!black, fill=#1!25, rounded corners=2pt, line width=0.6pt, inner sep=3pt, font=\fsbox, align=center},
  bxs/.style={draw=#1!85!black, fill=#1, rounded corners=2pt, line width=0.6pt, inner sep=3pt, font=\fsbox, text=white, align=center},
  plain/.style={draw=black!70, fill=white, rounded corners=2pt, line width=0.6pt, inner sep=3pt, font=\fsbox, align=center},
  flow/.style={->, >=stealth, line width=0.7pt, draw=black!75},
  stop/.style={->, >=stealth, line width=0.7pt, draw=softcoral},
  lab/.style={font=\fslab, inner sep=1pt, text=black!80},
  small/.style={font=\fslab, text=black!70, inner sep=1pt},
  chip/.style={anchor=north west, text width=2.3cm},
]
\definecolor{softcoral}{RGB}{242,133,108}
\definecolor{tealblue}{RGB}{54,144,192}
\definecolor{set2cyan}{RGB}{114,182,161}
\definecolor{set2orange}{RGB}{232,149,117}
\definecolor{cream}{RGB}{252,246,222}
\def\ytop{0} \def\ybot{-3.62} \def\ym{-0.98}

\draw[panel] (-0.2,\ytop)  rectangle (2.85,\ybot);   \node[head] at (1.325,-0.17) {Inputs};
\draw[panel] (3.25,\ytop)  rectangle (7.95,\ybot);   \node[head] at (5.6,-0.17)   {1.\ Preflight};
\draw[panel] (8.35,\ytop)  rectangle (10.4,\ybot);   \node[head] at (9.375,-0.17) {2.\ Synthesis};
\draw[panel] (10.85,\ytop) rectangle (14.65,\ybot);  \node[head] at (12.75,-0.17) {3.\ Certification};

\node[bx=set2orange, chip] (H)  at (0,-0.55) {baseline $H(\cdot)$};
\node[bxs=tealblue,  chip] (LB) at ($(H.south west)+(0,-1mm)$) {lower bound $\LB(\cdot)$};
\node[plain,         chip] (BR) at ($(LB.south west)+(0,-1mm)$) {level $\breq$};
\node[bx=set2cyan,   chip] (D)  at ($(BR.south west)+(0,-2.5mm)$) {instance\\ distribution $\sim\mathcal{D}$};
\coordinate (brk) at ($(H.north east)+(1.2mm,0)$);
\draw[black!55, line width=0.6pt] (brk) -- ++(0.8mm,0) |- ($(BR.south east)+(2mm,0)$) -- ++(-0.8mm,0);
\coordinate (brkmid) at ($(brk)!0.5!(brk |- BR.south)+(2mm,0)$);
\draw[flow] (brkmid) -- (brkmid -| 3.25,0);

\begin{scope}[shift={(3.5,-0.98)}]
  \foreach \i/\f in {0/1,1/0,2/0,3/1,4/0,5/0,6/0,7/1,8/0,9/0}{
    \ifnum\f=1 \fill[tealblue!70] (\i*2.1mm,0) rectangle ++(1.8mm,4mm);
    \else \draw[black!40] (\i*2.1mm,0) rectangle ++(1.8mm,4mm); \fi}
  \node[small, anchor=north west] at (0,-0.04) {filled: $H = \LB$};
\end{scope}
\node[plain, anchor=north west] (ceil) at (3.4,-1.55)
  {$\bceil=1-\Pr[H=\LB]$\\[2pt] {\fslab Is $\breq<\bceil$\,?}};
\draw[flow] (ceil.north |- 0,-1.32) -- (ceil.north);

\draw[flow] (ceil.east) -- (ceil.east -| 8.35,0);
\node[lab, anchor=south, yshift=0.4mm] at ($(ceil.east)!0.5!(ceil.east -| 7.95,0)$) {yes};

\node[draw=softcoral, fill=softcoral!12, rounded corners=2pt, line width=0.6pt,
      inner sep=3pt, font=\fslab, text=softcoral!85!black, align=center, anchor=north]
  (stop) at ($(ceil.south)+(0,-0.3)$) {ruled out\\ (unattainable)};
\draw[stop] (ceil.south) -- (stop.north)
  node[lab, midway, right=0.5mm, text=softcoral!85!black] {no};

\node[font=\fslab, align=center, text width=1.85cm, inner sep=0pt,
      text=black!85] at (9.375,-1.75)
  {e.g., Engram~\cite{karimi2026engram},\\
   ShinkaEvolve~\cite{shinkaevolve2025},\\
   \ldots\\[2pt]
   (\autoref{sec:eval-methods})};

\node[plain] (delta) at (12.75,\ym) {$\Delta(I) = H'(I) - H(I)$};
\draw[flow] (10.4,\ym) -- (delta.west);
\node[lab, anchor=south] at (10.625,\ym) [yshift=0.4mm] {$H'$};   
\begin{axis}[at={(11.15cm,-2.46cm)}, anchor=south west, scale only axis, width=2.5cm, height=0.95cm,
    xmin=0, xmax=8, ymin=0, ymax=1.15, axis lines=left,
    axis line style={black!45, line width=0.5pt, -{Stealth[length=1.5mm]}},
    xtick=\empty, ytick=\empty, clip=false,
    declare function={pdm(\x) = (\x)^1.8*exp(-\x/1.1)/0.5653;}]
  \addplot[draw=none, fill=softcoral!45, domain=4.5:7.5, samples=50] {pdm(x)} \closedcycle;
  \addplot[black!80, line width=0.9pt, smooth, domain=0.02:7.5, samples=100] {pdm(x)};
  \draw[black!60, densely dashed, line width=0.6pt] (axis cs:4.5,0) -- (axis cs:4.5,0.72);
  \draw[softcoral!90!black, line width=1.1pt] (axis cs:5.58,0) -- (axis cs:5.58,0.5);
  \node[small, anchor=west] at (axis cs:8.15,0) {$\Delta$};
  \coordinate (mtail) at (axis cs:6.6,0.42);
  \coordinate (mtop)  at (axis cs:4,1.15);
\end{axis}
\node[small, text=softcoral!85!black, anchor=south west, align=left] at ($(mtail)+(0.5mm,0.5mm)$) {worst\\[-1pt] $1-\breq$};
\draw[flow] (delta.south) -- (delta.south |- mtop);
\node[bxs=set2cyan, anchor=north] (cv) at (12.75,-2.72) {$\CVaR_{\breq}(\Delta)$};
\draw[flow] (12.75,-2.48) -- (cv.north);

\node[bx=cream, align=center, text width=1.62cm] (ver) at (cv.center -| 15.82,0)
  {verdict\\[1.5pt] {\fslab \textcolor{tealblue!85!black}{certifies}\\[-1pt]
                     \textcolor{black!60}{inconclusive}\\[-1pt]
                     \textcolor{softcoral!85!black}{positive bound}}};
\draw[flow] (cv.east) -- (ver.west);

\draw[flow] ($(D.south)+(0.35,0)$) |- (9.375,-3.92) -- (9.375,\ybot);
\node[lab, anchor=south] at (7.35,-3.92) {training set};
\draw[flow] ($(D.south)+(-0.35,0)$) |- (12.75,-4.22) -- (12.75,\ybot);
\draw[draw=softcoral, line width=1.1pt, dash pattern=on 2pt off 1.5pt] (3.05,-4.42) -- (3.05,-4.02);
\node[small, text=softcoral!85!black, anchor=north] at (3.05,-4.44) {firewall};
\node[lab, anchor=north west] at (3.42,-4.27) {$H$, $\breq$, and fresh certification instances $\sim\mathcal{D}$ (never shown to the agents)};
\end{tikzpicture}}
\caption{\textbf{\name end-to-end workflow.}
Given a baseline heuristic, a lower bound, a requested
tail level $\breq$, and a workload distribution, \name first checks
whether the certification target can be ruled out.
If not, an LLM-based coding agent generates candidate
heuristics using tail-focused feedback. The final candidate
is evaluated on independent instances, and its paired
differences with the baseline are used to determine the
certification verdict.}
\label{fig:cert}
\end{figure*}

\autoref{sec:motivation} showed how the synthesis objective
affects both heuristic performance and the strategies LLMs
discover. \name turns paired-CVaR feedback into a workflow
for synthesizing and certifying robust heuristics without
computing $\OPT$ (\autoref{fig:cert}). It consists of three
stages: checking certificate feasibility, synthesizing
candidate heuristics, and independently certifying
improvements on unseen workloads.

\subsection{Certifying Tail Improvement Without OPT}
\label{sec:tail-certificate}

Our goal is to reduce the tail of the optimality gap (\autoref{fig:motivation-tail}). Let $H$ denote the baseline heuristic, $H'$ a candidate replacement heuristic, and $\OPT(I)$ the optimal cost for workload $I \sim \mathcal{D}$. We express performance as a cost
to minimize, so $H(I)-\OPT(I)$ is the baseline's optimality gap.
Maximization problems are handled by negating the cost; all subsequent guarantees apply unchanged.
At tail level $\beta$, we want the worst $1-\beta$ fraction of the candidate's
gaps to be smaller, on average, than the worst $1-\beta$ fraction of the
baseline's gaps. Throughout the paper, the requested level $\breq$ is
$0.95$.

Directly comparing these gaps requires computing $\OPT$
over many large-scale workloads, which can be prohibitively expensive for combinatorial problems and impractical for repeated
LLM-based search (\autoref{app:opt}). Instead, we use the observable paired
difference $\Delta(I)=H'(I)-H(I)$, obtained by evaluating
both heuristics on the same workload.

\noindent\textbf{An observable certificate.}
Since $H'-\OPT=(H-\OPT)+\Delta$, subadditivity of
CVaR~\cite{artzner1999coherent,acerbi2002coherence} gives
\begin{equation}
\label{eq:certificate}
\begin{aligned}
&\CVaR_{\beta}(H'-\OPT)-\CVaR_{\beta}(H-\OPT)\\
&\hspace{3em}\leq \CVaR_{\beta}(\Delta).
\end{aligned}
\end{equation}
The left-hand side is the improvement we \emph{want} to establish; the
right-hand side is the quantity we can \emph{measure}. In particular, if
$\CVaR_{\beta}(\Delta)<0$, then the \textbf{candidate has a strictly smaller
tail optimality gap than the baseline} under $\mathcal{D}$, even though we
never computed $\OPT$
(proof in
\autoref{app:certificate-proof}). An analogous bound does not hold for quantiles.

Also, since the average of the worst
$1-\beta$ fraction is at least the overall average, we have:
\begin{equation}
\label{eq:mean-certificate}
\mathbb{E}[H'-H]\leq\CVaR_{\beta}(H'-H).
\end{equation}
Thus a \emph{negative paired CVaR also implies an average improvement over
the baseline}.

\begin{takeaway}
If $\CVaR_{\beta}(H'-H)$ is negative, we \textbf{certify}
improvements in both \textbf{tail robustness}
(at $\beta=0.95$, covering the worst 5\% tail)
and \textbf{average performance} relative
to the baseline.
\end{takeaway}

\subsection{Checking Headroom Before Synthesis}
\label{sec:limits-theory}
\label{sec:candidate-independent-limits}
\label{sec:preflight}

Before searching for a heuristic with negative paired CVaR,
we ask: \emph{can any feasible replacement heuristic achieve
that certificate?}
For example, suppose the baseline is already optimal on
10\% of workloads. No feasible candidate can improve on
the baseline for these workloads, so $H'(I) - H(I)\geq 0$
for at least 10\% of the distribution where $H'(I) \geq H(I)=\OPT(I)$.
Consequently, the worst 5\% of paired differences must
also be nonnegative, implying
$\CVaR_{0.95}(\Delta)\geq 0$ for every feasible candidate.
A negative certificate is therefore impossible at
$\beta=0.95$. In this example, $\beta$ must be below $0.9$.

\noindent\textbf{Detecting this limit without OPT.}
Determining where the baseline is optimal would itself require
solving for $\OPT$. We show that a valid lower bound $\LB(I)\leq\OPT(I)$ can establish baseline optimality without computing $\OPT$. For example, in VM packing, dividing total resource demand by per-machine capacity and rounding up yields a lower bound on the number of machines required. Whenever $H(I)=\LB(I)$, the baseline is optimal. The probability of this event therefore determines a candidate-independent \emph{certification ceiling}:

\begin{equation}
\label{eq:certification-ceiling}
\bceil=1-\Pr_{I\sim\mathcal{D}}[H(I)=\LB(I)].
\end{equation}
At requested tail $\breq\geq\bceil$, no feasible candidate
can achieve a strictly negative paired-CVaR certificate (\autoref{app:bounds}). 

\noindent\textbf{Preflight decision.}
\name first estimates the fraction of workloads on which
$H=\LB$ (\autoref{fig:cert}) from sampled workloads
and uses a one-sided confidence bound to rule out
certification only when the evidence is sufficient
(\autoref{app:bounds}). Otherwise, synthesis proceeds. Passing this check does \emph{not} guarantee that a
certifiable candidate exists; it only means that we have not ruled
one out. When a cheap valid lower bound is not available, preflight can be skipped.

\subsection{Using $\CVaR$ to Design Candidate Heuristics}
\label{sec:design-loop}

If the target has not been ruled out, \name uses an LLM-based harness
to propose candidate $H'$, executes it alongside the baseline on
training workloads, and returns the empirical
$\CVaR_{\breq}(H'-H)$ on the training data to the agent. Minimizing this score rewards
policies that improve the worst baseline-relative outcomes, rather
than allowing improvements elsewhere to conceal severe regressions (\autoref{sec:motivation}). Because candidates are assessed
repeatedly during synthesis, \name uses fewer training samples than the certification in the next step to keep training feedback affordable.

\subsection{From a Candidate to a Statistical Certificate}

After synthesis, we fix the selected candidate and evaluate
it alongside the baseline on independent held-out workloads.
Our theoretical guarantee requires the population paired CVaR
to be negative. However, observing a negative empirical CVaR
is not sufficient: the candidate might perform worse on
workloads we have not sampled.

We use the Dvoretzky--Kiefer--Wolfowitz (DKW)
inequality~\cite{dvoretzky1956asymptotic,massart1990tight}
for two purposes. First, it guides the number of independent
workloads needed for certification at a desired confidence level ($\delta$)
and quantile resolution. Second, it provides an upper
confidence bound $U_{\mathrm{DKW}}$ on the population paired
CVaR, accounting for sampling uncertainty (\autoref{app:statistical-procedure}).

For a fixed candidate, if $U_{\mathrm{DKW}}<0$, we certify
improvements in both mean performance and tail optimality
gaps with confidence at least $1-\delta$.
We use $\delta=0.01$, corresponding to 99\% confidence.
Otherwise, the result is inconclusive, even if the
empirical CVaR is negative.

\section{Evaluations}
\label{sec:more-cases}

We use Engram~\cite{karimi2026engram}
with Claude Code (Opus 5), a \$30 synthesis budget per run,
and five independent runs per configuration.
\name optimizes paired CVaR at $\beta=0.95$.
Final candidates are evaluated on held-out workloads
not used during synthesis. We use 99\% confidence for
statistical certification (\autoref{app:sizing}), approximately
28,000 independent workloads per benchmark. 
We express all problems as costs to minimize and define
$\Delta=H'-H$, where negative values indicate improvements
over the baseline. We report normalized mean cost improvement
as $-\mathbb{E}[\Delta]/|\mathbb{E}[H]|$ and normalized paired CVaR as $\CVaR_{0.95}(\Delta)/|\mathbb{E}[H]|$,
both expressed as percentages.
Higher mean cost improvement indicates better average performance,
while lower paired CVaR indicates better tail performance.
In particular, negative paired CVaR means that even the
worst 5\% of paired differences are negative on average.

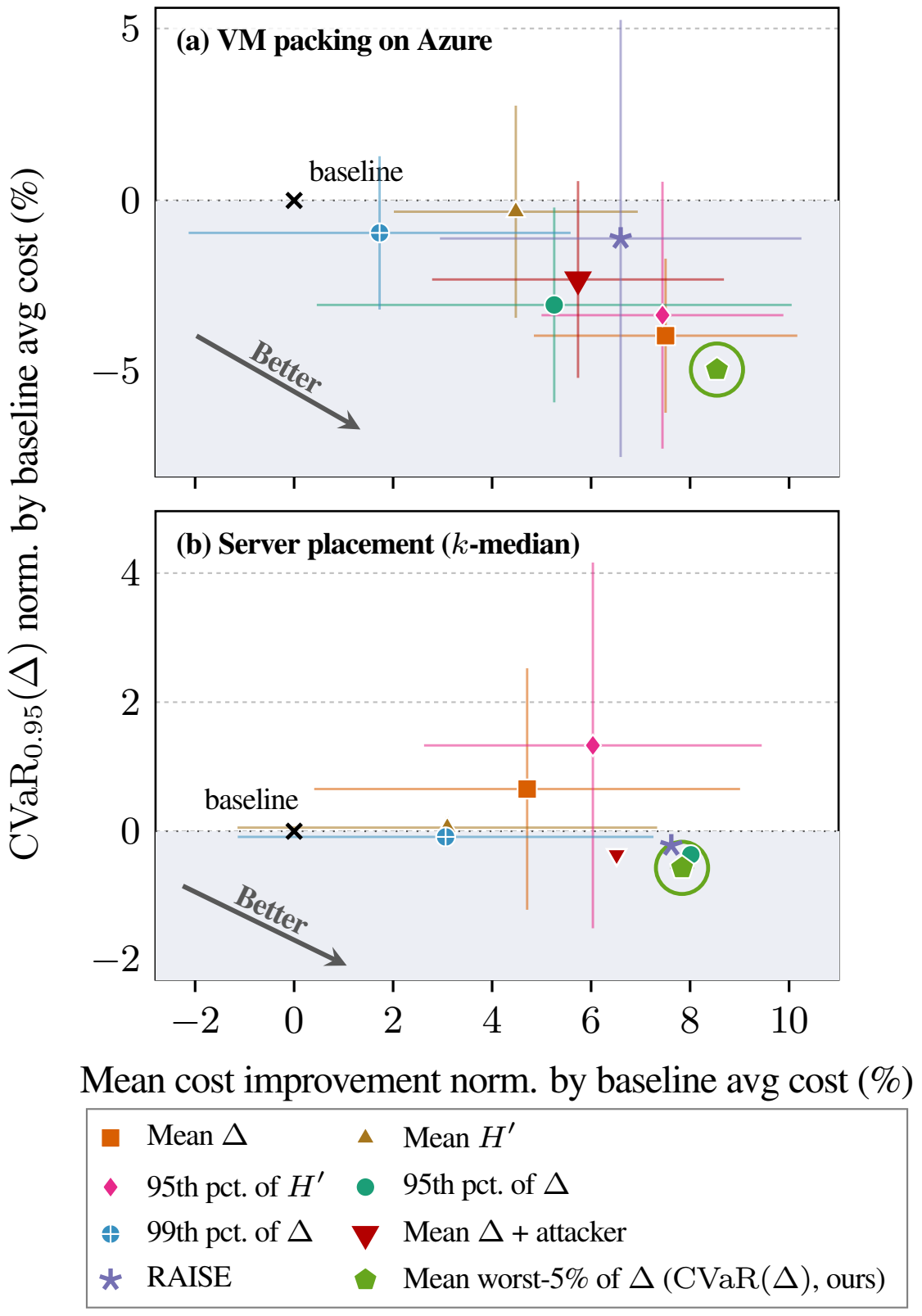
\begin{figure}[t]
\centering
\pgfplotsset{
  mk/.style={only marks, mark size=2.2pt,
    mark options={solid, draw=white, line width=0.4pt}},
}
\tikzset{
  eb/.style={line width=0.55pt, opacity=0.55},
  ttl/.style={font=\scriptsize\bfseries, fill=white, fill opacity=0.8,
    text opacity=1, inner sep=1.5pt},
}

\begin{tikzpicture}
\begin{groupplot}[
  group style={group size=1 by 2, vertical sep=8pt,
               xlabels at=edge bottom, xticklabels at=edge bottom},
  paperaxis, width=0.86\linewidth, height=0.65\linewidth,
  xmin=-2.8, xmax=11,
  xlabel={Mean cost improvement norm. by baseline avg cost (\%)},
  /tikz/eb/.style={line width=0.55pt, opacity=0.55},
  mk/.style={only marks, mark size=2.2pt, mark options={solid, draw=white, line width=0.4pt}},
  /tikz/ttl/.style={font=\scriptsize\bfseries, fill=white, fill opacity=0.8, text opacity=1, inner sep=1.5pt}]

\nextgroupplot[ymin=-8.02, ymax=5.60,
  legend to name=morelegend, legend columns=2,
  legend style={font=\scriptsize, /tikz/column sep=4pt}]
\fill[set2blue!18] (axis cs:-2.8,-8.02) rectangle (axis cs:11,0);
\draw[set2gray, dotted] (axis cs:-2.8,0) -- (axis cs:11,0);
\addplot[only marks, mark=x, mark size=2.4pt, color=black, line width=0.9pt, forget plot] coordinates {(0,0)};
\node[font=\scriptsize, anchor=south west] at (axis cs:0.,0.1) {baseline};
\draw[dark2orange, eb] (axis cs:4.844,-3.931) -- (axis cs:10.166,-3.931) (axis cs:7.505,-6.171) -- (axis cs:7.505,-1.691);
\addplot[mk, mark=square*, color=dark2orange] coordinates {(7.505,-3.931)};
\addlegendentry{Mean $\Delta$}
\draw[dark2brown, eb] (axis cs:2.015,-0.327) -- (axis cs:6.943,-0.327) (axis cs:4.479,-3.411) -- (axis cs:4.479,2.757);
\addplot[mk, mark=triangle*, mark size=2.5pt, color=dark2brown] coordinates {(4.479,-0.327)};
\addlegendentry{Mean $H'$}
\draw[dark2pink, eb] (axis cs:4.998,-3.335) -- (axis cs:9.885,-3.335) (axis cs:7.442,-7.215) -- (axis cs:7.442,0.545);
\addplot[mk, mark=diamond*, mark size=2.5pt, color=dark2pink] coordinates {(7.442,-3.335)};
\addlegendentry{95th pct.\ of $H'$}
\draw[dark2teal, eb] (axis cs:0.458,-3.037) -- (axis cs:10.052,-3.037) (axis cs:5.255,-5.869) -- (axis cs:5.255,-0.205);
\addplot[mk, mark=*, color=dark2teal] coordinates {(5.255,-3.037)};
\addlegendentry{95th pct.\ of $\Delta$}
\draw[tealblue, eb] (axis cs:-2.131,-0.942) -- (axis cs:5.586,-0.942) (axis cs:1.727,-3.170) -- (axis cs:1.727,1.286);
\addplot[mk, mark=oplus*, color=tealblue] coordinates {(1.727,-0.942)};
\addlegendentry{99th pct.\ of $\Delta$}
\draw[red!70!black, eb] (axis cs:2.788,-2.297) -- (axis cs:8.684,-2.297) (axis cs:5.736,-5.157) -- (axis cs:5.736,0.563);
\addplot[mk, mark=triangle*, mark size=2.5pt, color=red!70!black, mark options={rotate=180}] coordinates {(5.736,-2.297)};
\addlegendentry{Mean $\Delta$ + attacker}
\draw[azpurple, eb] (axis cs:2.945,-1.105) -- (axis cs:10.249,-1.105) (axis cs:6.597,-7.458) -- (axis cs:6.597,5.248);
\addplot[only marks, mark=star, mark size=2.6pt, color=azpurple, mark options={solid, line width=1pt}] coordinates {(6.597,-1.105)};
\addlegendentry{RAISE}
\draw[dark2green, eb] (axis cs:8.381,-4.916) -- (axis cs:8.705,-4.916) (axis cs:8.543,-5.056) -- (axis cs:8.543,-4.777);
\addplot[mk, mark=pentagon*, mark size=2.9pt, color=dark2green] coordinates {(8.543,-4.916)};
\addlegendentry{Mean worst-5\% of $\Delta$ ($\CVaR(\Delta)$, ours)}
\draw[dark2green, line width=0.9pt] (axis cs:8.543,-4.916) circle (6pt);
\betterD{0.06}{0.30}{0.30}{0.10}
\node[ttl, anchor=north west] at (rel axis cs:0.02,0.97) {(a) VM packing on Azure};

\nextgroupplot[ymin=-2.30, ymax=4.96]
\fill[set2blue!18] (axis cs:-2.8,-2.30) rectangle (axis cs:11,0);
\draw[set2gray, dotted] (axis cs:-2.8,0) -- (axis cs:11,0);
\addplot[only marks, mark=x, mark size=2.4pt, color=black, line width=0.9pt, forget plot] coordinates {(0,0)};
\node[font=\scriptsize, anchor=south east] at (axis cs:0.4,0.1) {baseline};
\draw[dark2orange, eb] (axis cs:0.406,0.654) -- (axis cs:9.010,0.654) (axis cs:4.708,-1.217) -- (axis cs:4.708,2.525);
\addplot[mk, mark=square*, color=dark2orange, forget plot] coordinates {(4.708,0.654)};
\draw[dark2brown, eb] (axis cs:-1.144,0.056) -- (axis cs:7.334,0.056) (axis cs:3.095,-0.113) -- (axis cs:3.095,0.225);
\addplot[mk, mark=triangle*, mark size=2.5pt, color=dark2brown, forget plot] coordinates {(3.095,0.056)};
\draw[dark2pink, eb] (axis cs:2.625,1.330) -- (axis cs:9.447,1.330) (axis cs:6.036,-1.504) -- (axis cs:6.036,4.164);
\addplot[mk, mark=diamond*, mark size=2.5pt, color=dark2pink, forget plot] coordinates {(6.036,1.330)};
\draw[dark2teal, eb] (axis cs:8.007,-0.365) -- (axis cs:8.023,-0.365) (axis cs:8.015,-0.368) -- (axis cs:8.015,-0.362);
\addplot[mk, mark=*, color=dark2teal, forget plot] coordinates {(8.015,-0.365)};
\draw[tealblue, eb] (axis cs:-1.139,-0.088) -- (axis cs:7.260,-0.088) (axis cs:3.061,-0.248) -- (axis cs:3.061,0.072);
\addplot[mk, mark=oplus*, color=tealblue, forget plot] coordinates {(3.061,-0.088)};
\draw[dark2green, eb]
  (axis cs:7.596,-0.568) -- (axis cs:8.079,-0.568)
  (axis cs:7.837,-0.701) -- (axis cs:7.837,-0.435);
\addplot[mk, mark=pentagon*, mark size=2.9pt, color=dark2green, forget plot] coordinates {(7.837,-0.568)};
\draw[dark2green, line width=0.9pt] (axis cs:7.837,-0.568) circle (6pt);
\draw[azpurple, eb]
  (axis cs:7.571,-0.220) -- (axis cs:7.669,-0.220)
  (axis cs:7.620,-0.240) -- (axis cs:7.620,-0.200);
\addplot[only marks, mark=star, mark size=2.6pt, color=azpurple, mark options={solid, line width=1pt}, forget plot] coordinates {(7.620,-0.220)};

\draw[red!70!black, line width=0.9pt]
  (axis cs:6.511,-0.365) -- (axis cs:6.521,-0.365)
  (axis cs:6.516,-0.369) -- (axis cs:6.516,-0.361);
\addplot[only marks, mark=triangle*, mark size=2.6pt, color=red!70!black, mark options={solid, rotate=180, draw=white, line width=0.5pt}, forget plot] coordinates {(6.516,-0.365)};

\betterD{0.04}{0.20}{0.28}{0.03}
\node[ttl, anchor=north west] at (rel axis cs:0.02,0.97) {(b) Server placement ($k$-median)};
\end{groupplot}

\node[rotate=90, anchor=south, font=\footnotesize, align=center]
  at ($(group c1r1.outer south west)!0.5!(group c1r2.outer north west)+(-2pt,0)$)
  {$\CVaR_{0.95}(\Delta)$ norm. by baseline avg cost (\%)};
\node[anchor=north] at ($(group c1r2.south)+(0,-26pt)$) {\pgfplotslegendfromname{morelegend}};
\end{tikzpicture}
\caption{\textbf{Paired-CVaR objective achieves strong mean--tail trade-offs.} Normalized mean cost improvement versus empirical
paired CVaR at $\beta=0.95$ for (a) Azure VM packing and (b) server placement ($k$-median). Both metrics are expressed as percentages of the
baseline's average cost magnitude.
Markers show means over five independent synthesis runs, with standard-deviation bars. Higher mean improvement and lower paired CVaR
indicate better performance.}
\vspace{-15 pt}
\label{fig:more-gain-cvar}
\end{figure}

\subsection{Case-Study: Azure VM Packing}
\label{sec:case-vm}

We study online VM packing using production Azure
traces~\cite{azurepackingtrace2020,hadary2020protean}.
VMs arrive sequentially and must be assigned to
machines with sufficient CPU and memory capacity,
aiming to minimize the number of machines used.
The baseline is tight packing (Kubernetes
``MostAllocated'')~\cite{k8s-noderesourcesfit},
which places each VM on the machine it fills most.
We use a capacity-based lower bound on the optimum.

\autoref{fig:more-gain-cvar}(a) shows that paired-CVaR
achieves the largest mean cost improvement and lowest empirical
$\CVaR_{0.95}(\Delta)$ among the evaluated objectives.
Compared with mean-optimized heuristics, it achieves
14\% larger mean cost improvements and reduces empirical
paired CVaR from $-3.93\%$ to $-4.92\%$,
a 25\% larger tail improvement.
Compared with optimizing the 95th percentile of
$\Delta$, paired-CVaR increases mean cost improvement from
5.26\% to 8.54\% and reduces empirical
$\CVaR_{0.95}(\Delta)$ from $-3.04\%$ to $-4.92\%$.
Paired-CVaR is also the only objective that
certifies all five synthesis runs, with no observed
regressions and the lowest run-to-run variability.

\subsection{Case-Study: Server Placement ($k$-Median)}
\label{sec:case-kmedian}

We consider $k$-median server placement, where $k$
server locations are selected to minimize total
client--server distance~\cite{qiu2001placement}.
Workloads include spatially dispersed and clustered
clients~\cite{bhattacharya2014robust}. The baseline
greedily selects locations that minimize service cost,
and we use an LP relaxation as a lower bound.
Unlike online admission control, all client locations
are known, allowing us to test robustness under
spatial workload variation rather than uncertain arrivals.

As shown in \autoref{fig:more-gain-cvar}(b),
paired-CVaR achieves 66\% larger mean cost improvements
than mean-optimized heuristics (7.84\% vs.\
4.71\%), while changing empirical
$\CVaR_{0.95}(\Delta)$ from $+0.65\%$ to
$-0.57\%$. Mean-optimized heuristics
have positive paired CVaR, while
\name-generated heuristics achieve negative
paired CVaR.
All five runs have negative empirical paired CVaR,
and three of five are statistically certified
(\autoref{app:uncertainty}).
In contrast, optimizing either the 99th percentile
of $\Delta$ or the 95th percentile of $H'$
certifies only two of five runs.

\begin{takeaway}
Across these case studies
(\autoref{sec:motivation}, \autoref{sec:more-cases}),
paired-CVaR offers the strongest mean--tail
trade-off among the evaluated synthesis objectives.
Paired-CVaR achieves negative empirical
$\CVaR_{0.95}(\Delta)$ across all three problems and retains substantial mean improvements.
It certifies all five synthesis runs on both Azure
benchmarks and three of five on Server Placement.
\end{takeaway}

\section{Ablation Studies}
\label{sec:eval}

The case studies show how synthesis objective
affects heuristic performance and the strategies LLMs
discover. We now evaluate if these improvements generalize across other benchmarks,
how they compare with worst-case feedback and human-designed
heuristics, and how certification depends on model capability and search architecture. Unless noted, we use the setup from~\autoref{sec:more-cases}. Workload, complexity, and lower-bound details are
in~\autoref{app:eval_detail}.

\noindent \textbf{Benchmarks}\label{sec:eval-setup}
We evaluate five additional benchmarks; Makespan is ruled out at preflight, leaving seven for synthesis.

\emph{Online Weighted Matching.}
Weighted edges arrive sequentially on a graph and must be accepted or
rejected irrevocably, with no two accepted edges sharing
an endpoint~\cite{karp1990optimal,mehta2013online}. The goal is to maximize the total weight of accepted edges.
This models throughput allocation in datacenter networks
~\cite{mwm-tub,mwm-understanding}. The baseline heuristic accepts
an edge whenever both endpoints are available.

\emph{Two Dimensional Vector Bin Packing (VBP2D).}
We study offline and online resource packing, aiming to use as few machines as possible to meet CPU and memory demands~\cite{garey1976resource,johnson1974worst,
panigrahy2011heuristics}. The baseline for both is First Fit. These workloads differ
from the Azure VM Packing and are synthetic workloads (\autoref{app:workflow}).

\emph{Traffic Engineering (TE MLU).}
We route traffic to minimize maximum
link utilization~\cite{fortz2000internet,hong2013swan,
jain2013b4,bogle2019teavar}, using shortest-path routing
as the baseline. We evaluate SWAN, Abilene, B4, Uninet,
and Cogentco topologies~\cite{hong2013swan,jain2013b4,
knight2011zoo}.

\emph{Makespan.}
We assign jobs with processing times to
identical machines, aiming to minimize the makespan~\cite{graham1969bounds}.
The baseline is
Longest Processing Time first (LPT), which assigns
jobs in decreasing order of processing time to the
least-loaded machine.
This problem is ruled out in the preflight phase at requested certificate level because LPT matches the lower bound on 35.5\% of workloads (\autoref{sec:eval-preflight}). 

\subsection{\name Finds Certifiable Improvements
Across Diverse Problems}
\label{sec:eval-generalization}

\begin{figure}
\centering
\definecolor{azpurple}{RGB}{117,112,179}
\begin{tikzpicture}
\begin{axis}[paperaxis, width=0.82\linewidth, height=0.74\linewidth,
  ylabel={Mean worst-5\% of $\Delta$\\norm.\ by baseline's avg cost (\%)},
  xlabel={Mean cost improvement over $H$\\norm.\ by baseline's avg cost (\%)},
  xlabel style={align=center}, ylabel style={align=center},
  xmin=-2, xmax=29, ymin=-7, ymax=1.6,
  xtick={0,5,10,15,20,25}, ytick={-6,-4,-2,0}, yticklabels={$-6$,$-4$,$-2$,$0$},
  clip=false,
  ebars/.style={error bars/x dir=both, error bars/x explicit,
    error bars/y dir=both, error bars/y explicit,
    error bars/error bar style={color=#1, line width=0.6pt}},
  pt/.style={only marks, mark size=2.0pt, color=#1,
    mark options={solid, fill=#1, draw=white, line width=0.4pt}, ebars=#1},
  /tikz/lb/.style={font=\scriptsize, inner sep=1pt},
  /tikz/lead/.style={black!35, line width=0.4pt}]
\fill[set2blue!18] (axis cs:-2,-7) rectangle (axis cs:29,0);
\draw[set2gray, dotted] (axis cs:-2,0) -- (axis cs:29,0);
\draw[set2gray, dotted] (axis cs:0,-7) -- (axis cs:0,1.6);

\addplot[mark=*, pt=tealblue]    coordinates {(11.13,-0.78) +- (0.02,0.14)};  
\addplot[mark=*, pt=dark2green]  coordinates {(6.51,-0.43) +- (0.06,0.04)};   
\addplot[mark=*, pt=azpurple]    coordinates {(26.11,-0.63) +- (1.10,0.30)};  
\addplot[only marks, mark=o, mark size=2.2pt, color=set2gray!70!black, line width=0.9pt,
  ebars=set2gray!70!black] coordinates {(0.98,0.04) +- (1.18,0.08)};          

\addplot[mark=square*, pt=dark2orange] coordinates {(7.84,-0.27) +- (0.24,0.13)};  
\addplot[mark=square*, pt=dark2pink]   coordinates {(12.31,-5.98) +- (0.85,0.34)}; 
\addplot[mark=square*, pt=dark2brown]  coordinates {(8.54,-4.92) +- (0.16,0.14)};  

\node[lb, text=set2gray!70!black, anchor=south west]
  at (axis cs:-1.6,0.35) {Online VBP2D (0/5)};

\node[lb, text=dark2orange!85!black, anchor=south west]
  at (axis cs:8.2,-0.5) {Server Placement (3/5)};

\draw[lead] (axis cs:6.4,-0.6) -- (axis cs:6.4,-1.45);
\node[lb, text=dark2green!85!black, anchor=north east]
  at (axis cs:6.4,-1.45) {VBP2D (5/5)};

\draw[lead] (axis cs:11.13,-0.98) -- (axis cs:11.13,-1.85);
\node[lb, text=tealblue!85!black, anchor=north west]
  at (axis cs:10.5,-1.85) {Online wtd.\ matching (2/5)};

\node[lb, text=azpurple!85!black, anchor=north]
  at (axis cs:24,-1.1) {TE MLU (4/5)};

\node[lb, text=dark2brown!85!black, anchor=west]
  at (axis cs:9.3,-4.92) {Azure VM packing (5/5)};

\node[lb, text=dark2pink!85!black, anchor=west]
  at (axis cs:11.4,-6.7) {GPU Batch Admission (5/5)};
  
\end{axis}
\end{tikzpicture}
\caption{\textbf{\name achieves mean and tail improvements
across six of seven benchmarks.}
Mean cost improvement versus empirical paired-CVaR on held-out workloads,
normalized per problem by $|\mathbb{E}[H]|$, the baseline's
average cost on the certification set.
Markers show mean $\pm$ standard deviation over five
independent synthesis runs. Labels indicate the number of
certified runs out of five
(\autoref{tab:dkw}).}
\vspace{-10 pt}
\label{fig:eval-cert-rate}
\end{figure}
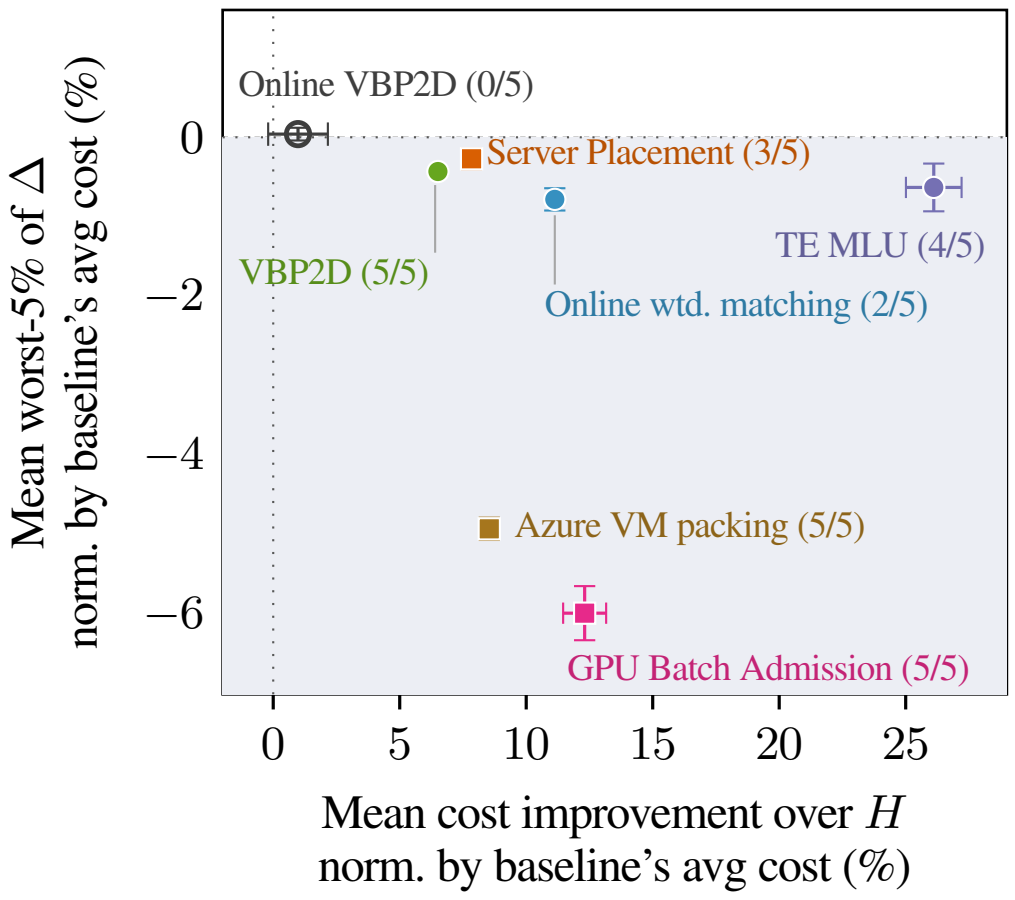

Across seven different benchmarks, \name certifies
improvements in mean performance and tail optimality
gaps in 24 of 35 independent synthesis runs
(\autoref{fig:eval-cert-rate}). All five runs are certified on Azure GPU Batch Admission (\autoref{sec:motivation}), Azure VM Packing, and Offline VBP2D. Certification also succeeds
in four of five runs on TE MLU, three of five on server
placement, and two of five on Online Weighted Matching.

In Online VBP2D, candidates improve average performance, but none of the five runs certifies a tail improvement. Even though preflight does not rule out headroom (\autoref{sec:eval-preflight}), no tested model (GPT-4o-mini, Sonnet 5.5, Opus 5), training objective (\autoref{sec:motivation}), or synthesis harness~\cite{karimi2026engram,karimi2025robusta} achieves certification (\autoref{sec:eval-methods}).
Optimizing a tail-aware objective does not guarantee that search will find a certifiably robust heuristic, nor does passing preflight establish that one is reachable. So, \textsc{Tailor}'s post-synthesis certification remains essential to catch these cases and prevent operators from deploying heuristics with unverified tail risk.

\subsection{Small-Scale Worst-Case Optimization Does Not Ensure Tail Robustness}
\label{sec:maxgap}

In this section, we ask whether a heuristic optimized to
directly reduce the optimality gap on small-scale instances,
where computing $\OPT$ is affordable, is also robust at
larger scales. We use MetaEase~\cite{karimi2026metaease} to
discover large optimality gaps on these small-scale workloads
and use the largest discovered gap as synthesis feedback.
We compare this with distributional paired-CVaR feedback,
using the same Opus 5 model and Engram harness.
Both approaches are evaluated on the same larger,
held-out workload distributions.

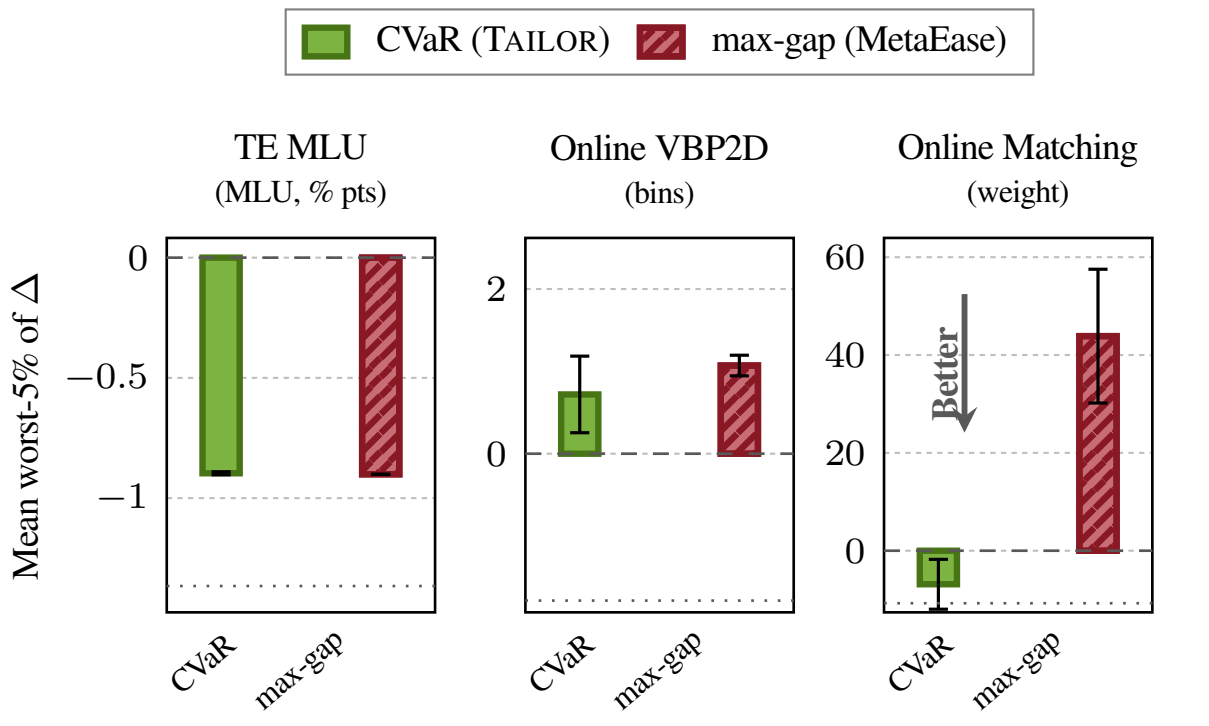
\begin{figure}[t]
\centering
\begin{tikzpicture}
\definecolor{maxgapred}{RGB}{170,30,45}
\begin{groupplot}[group style={group size=3 by 1, horizontal sep=17pt}, barplot/axis, scale only axis, width=0.215\linewidth, height=0.30\linewidth,
  /pgf/bar width=7pt, enlarge x limits=0.7, symbolic x coords={CVaR,max-gap}, xtick={CVaR,max-gap},
  xticklabel style={font=\tiny, rotate=40, anchor=east}, yticklabel style={font=\scriptsize}, ylabel style={font=\scriptsize, yshift=-2pt},
  title style={font=\scriptsize, align=center, yshift=-1pt},
  error bars/y dir=both, error bars/y explicit, error bars/error bar style={line width=0.6pt, black},
  error bars/error mark options={rotate=90, mark size=1.8pt, line width=0.6pt, black}]
\nextgroupplot[title={TE MLU\\{\tiny (MLU, \% pts)}}, ymin=-1.475, ymax=0.082, ylabel={Mean worst-5\% of $\Delta$}, legend to name=maxgaplegend, legend columns=2, legend style={font=\scriptsize, /tikz/column sep=3pt}]
\addplot[draw=dark2green!70!black, fill=dark2green!85] coordinates {(CVaR,-0.8967) +- (0,0.0062)};
\addlegendentry{CVaR (\name)}
\addplot[draw=maxgapred!80!black, fill=maxgapred!65, postaction={pattern={Lines[angle=45,distance=2.5pt]}, pattern color=maxgapred!80!black}] coordinates {(max-gap,-0.9009) +- (0,0.0004)};
\addlegendentry{max-gap (MetaEase)}
\draw[set2gray, line width=0.5pt, dashed] ({rel axis cs:0,0}|-{axis cs:CVaR,0}) -- ({rel axis cs:1,0}|-{axis cs:CVaR,0});
\draw[set2gray, line width=0.5pt, dotted] ({rel axis cs:0,0}|-{axis cs:CVaR,-1.3656}) -- ({rel axis cs:1,0}|-{axis cs:CVaR,-1.3656});
\nextgroupplot[title={Online VBP2D\\{\tiny (bins)}}, ymin=-1.926, ymax=2.620]
\addplot[draw=dark2green!70!black, fill=dark2green!85] coordinates {(CVaR,0.7198) +- (0,0.4650)};
\addplot[draw=maxgapred!80!black, fill=maxgapred!65, postaction={pattern={Lines[angle=45,distance=2.5pt]}, pattern color=maxgapred!80!black}] coordinates {(max-gap,1.0708) +- (0,0.1252)};
\draw[set2gray, line width=0.5pt, dashed] ({rel axis cs:0,0}|-{axis cs:CVaR,0}) -- ({rel axis cs:1,0}|-{axis cs:CVaR,0});
\draw[set2gray, line width=0.5pt, dotted] ({rel axis cs:0,0}|-{axis cs:CVaR,-1.7833}) -- ({rel axis cs:1,0}|-{axis cs:CVaR,-1.7833});
\nextgroupplot[title={Online Matching\\{\tiny (weight)}}, ymin=-12.597, ymax=63.938]
\addplot[draw=dark2green!70!black, fill=dark2green!85] coordinates {(CVaR,-6.8582) +- (0,5.0948)};
\addplot[draw=maxgapred!80!black, fill=maxgapred!65, postaction={pattern={Lines[angle=45,distance=2.5pt]}, pattern color=maxgapred!80!black}] coordinates {(max-gap,43.8653) +- (0,13.6744)};
\draw[set2gray, line width=0.5pt, dashed] ({rel axis cs:0,0}|-{axis cs:CVaR,0}) -- ({rel axis cs:1,0}|-{axis cs:CVaR,0});
\draw[set2gray, line width=0.5pt, dotted] ({rel axis cs:0,0}|-{axis cs:CVaR,-10.7342}) -- ({rel axis cs:1,0}|-{axis cs:CVaR,-10.7342});
\betterV{0.3}{0.85}{0.48}
\end{groupplot}
\node[anchor=south] at ($(group c1r1.north east)!0.5!(group c3r1.north west)+(0,27pt)$) {\pgfplotslegendfromname{maxgaplegend}};
\end{tikzpicture}
\caption{\textbf{Small-scale worst-case feedback does not
reliably transfer to large-scale tail robustness.}
Empirical paired-CVaR on large, held-out workloads
after synthesis using distributional paired-CVaR or
MetaEase-based small-scale maximum-gap feedback.
Values are in each benchmark's native units;
lower is better. Error bars show standard deviations
across five independent synthesis runs.}
\vspace{-10 pt}
\label{fig:maxgap}
\end{figure}

\autoref{fig:maxgap} shows that reducing worst-case
optimality gaps on small instances does not necessarily
improve tail robustness on larger workloads.
The clearest difference is on Online Weighted Matching:
paired-CVaR synthesis achieves negative empirical
paired CVaR, while maximum-gap synthesis produces
substantial tail regressions.
The two perform similarly on TE MLU.
On Online VBP2D, neither achieves negative empirical
paired CVaR on average, but maximum-gap heuristics
regress on 11\% of held-out workloads, compared with
0.08\% for paired-CVaR heuristics.

\subsection{Generated Heuristics Outperform Human Controls}
\label{sec:eval-best}
We evaluate the strength of the certified improvements by comparing generated heuristics with
strong human-designed alternatives, and by measuring how
conservative the certificate is relative to the true
change in tail optimality gaps.

\begin{figure}[t]
\centering
\definecolor{capsyn}{RGB}{94,79,162}
\definecolor{capctl}{RGB}{255,217,47}
\vspace{3pt}
\begin{tikzpicture}
\begin{axis}[paperaxis, xbar, scale only axis,
  width=0.63\linewidth, height=4.2cm, bar width=6pt,
  xmin=-112, xmax=112, ymin=-0.5, ymax=5.5, y dir=reverse,
  xtick={-100,-50,0,50,100},
  xticklabels={$-100$,$-50$,$0$,$+50$,$+100$},
  ytick={0,1,2,3,4,5},
  yticklabels={Server Placement,GPU Batch Admission,Online matching,VBP2D,TE MLU,Azure VM Packing},
  tick label style={font=\scriptsize}, yticklabel style={align=right},
  xlabel={Normalized paired-CVaR (\%); lower is better},
  label style={font=\scriptsize},
  axis x line*=bottom, axis y line*=left, y axis line style={draw=none},
  ymajorgrids=false, xmajorgrids=true, tick style={draw=none},
  clip=false,
  legend image code/.code={\draw[#1] (0pt,-0.6ex) rectangle (0.8em,0.6ex);},
  legend columns=2, legend cell align=left,
  legend style={font=\scriptsize, at={(0.35,1.04)}, anchor=south,
    draw=none, fill=none, /tikz/column sep=5pt}]
\draw[black!55, line width=0.6pt] (axis cs:0,-0.5) -- (axis cs:0,5.5);
\addplot[bar shift=3.5pt, fill=capsyn, draw=capsyn, line width=0.4pt]
  coordinates {(-96.45,0) (-79.19,1) (-99.55,2) (-30.17,3) (-66.01,4) (-53.2,5)};
\addlegendentry{\name}
\addplot[bar shift=-3.5pt, fill=capctl, draw=capctl!65!black, line width=0.4pt]
  coordinates {(-88.13,0) (-29.18,1) (-37.44,2) (-24.08,3) (97.55,4) (18.6,5)};
\addlegendentry{Human}
\foreach \x/\y/\v in {-96.45/0/-96.5,-79.19/1/-79.2,-99.55/2/-99.5,-66.01/4/-66.0,-53.2/5/-53.2}{
  \edef\temp{\noexpand\node[font=\noexpand\scriptsize, text=white, anchor=west,
    xshift=2pt, yshift=3.5pt, inner sep=0pt] at (axis cs:\x,\y) {$\v$};}\temp
}
\foreach \x/\y/\v in {-88.13/0/-88.1,-37.44/2/-37.4}{
  \edef\temp{\noexpand\node[font=\noexpand\scriptsize, anchor=west,
    xshift=2pt, yshift=-3.5pt, inner sep=0pt] at (axis cs:\x,\y) {$\v$};}\temp
}
\node[font=\scriptsize, anchor=east, xshift=-2pt, yshift=3.5pt, inner sep=0pt]
  at (axis cs:-30.17,3) {$-30.2$};
\node[font=\scriptsize, anchor=east, xshift=-2pt, yshift=-3.5pt, inner sep=0pt]
  at (axis cs:-24.08,3) {$-24.1$};
\node[font=\scriptsize, anchor=east, xshift=-2pt, yshift=-3.5pt, inner sep=0pt]
  at (axis cs:-29.18,1) {$-29.2$};
\node[font=\scriptsize, anchor=east, xshift=-2pt, yshift=-3.5pt, inner sep=0pt]
  at (axis cs:97.55,4) {$+97.6$};
\node[font=\scriptsize, anchor=west, xshift=2pt, yshift=-3.5pt, inner sep=0pt]
  at (axis cs:18.6,5) {$+18.6$};
\node[font=\scriptsize, text=rejline, anchor=west, xshift=2pt, yshift=5pt, inner sep=0pt]
  at (axis cs:0,4) {positive certificate};
\node[font=\scriptsize, text=rejline, anchor=west, xshift=2pt, yshift=5pt, inner sep=0pt]
  at (axis cs:0,5) {positive certificate};
\betterH{0.80}{0.93}{0.63}
\end{axis}
\end{tikzpicture}\par
\caption{\textbf{Best observed \name-generated heuristics
versus human-designed controls.}
Empirical paired-CVaR relative to the same baseline,
normalized by the magnitude of the benchmark-specific
empirical floor $F_{0.95}<0$
(\autoref{app:floor-ceiling}).
Lower is better; $-100\%$ reaches the theoretical
floor, which need not be attainable.}
\vspace{-10 pt}
\label{fig:captured}
\end{figure}
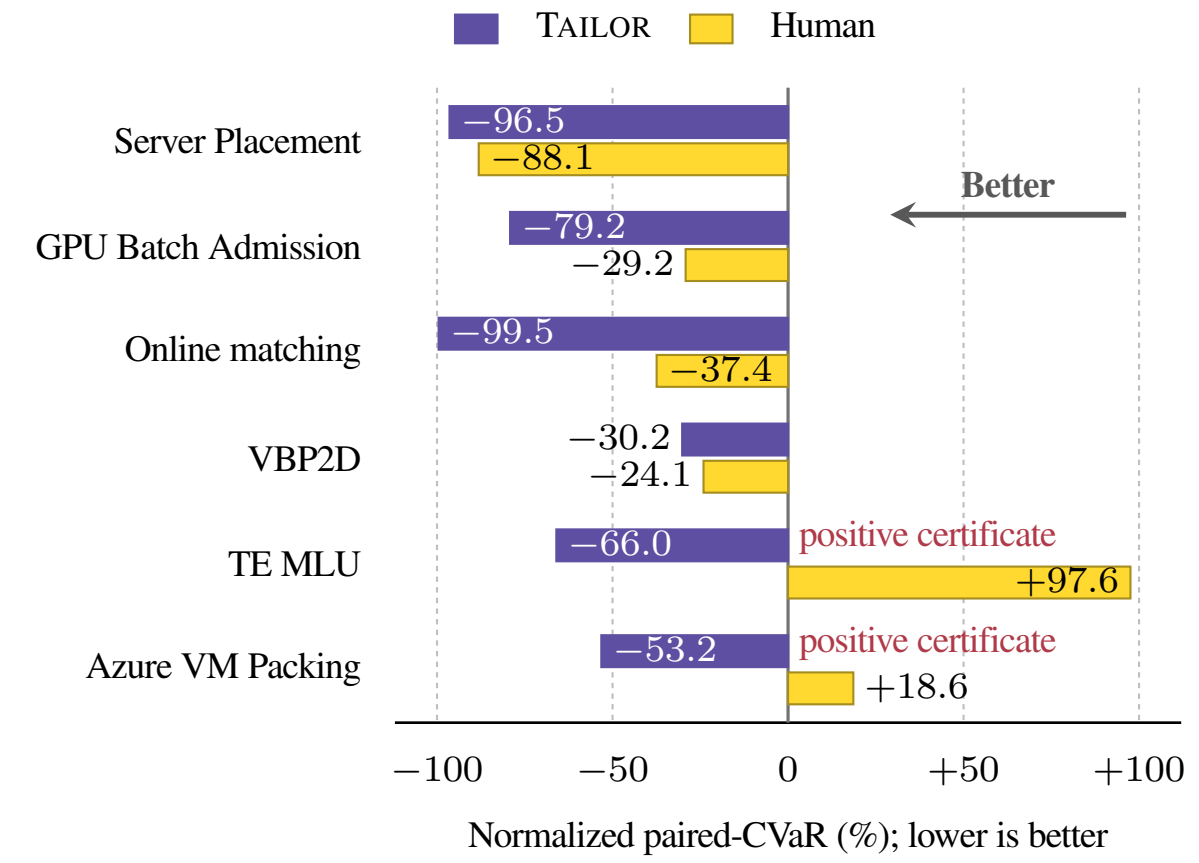

\noindent\textbf{Comparison with human-designed heuristics.}
We compare the best heuristics generated by \name
with established human-designed controls
(\autoref{tab:eval-benchmarks}).
\autoref{fig:captured} reports their empirical paired-CVaR
relative to the same baseline and workload distribution.
Across all six benchmarks, \name
achieves a lower paired-CVaR than the human control.

Since benchmark costs have different units, we normalize
each value by the magnitude of the certificate floor
$F_{0.95}=\CVaR_{0.95}(\LB-H)$
(\autoref{app:floor-ceiling}).
This is a candidate-independent lower bound on achievable
paired-CVaR. A normalized value of $-100\%$ reaches this
theoretical floor, which does not need to be attainable. \name's generated heuristics for Server Placement
and Online Weighted Matching approach this bound.
The best generated heuristics also match or outperform
human-designed controls in average cost improvement across all six
benchmarks. The largest differences are on GPU Batch
Admission (12.2\% vs.\ 6.2\%) and Azure VM Packing
(9.0\% vs.\ 0.1\%).

\noindent\textbf{Relation to true optimality gaps.}
To assess the tightness of the bound in
\autoref{eq:certificate}, we solve 28,000 small-scale
$k$-median instances to optimality
(\autoref{app:opt})---enough that the empirical CDF of
optimality gaps is resolved to within a DKW error of about
$0.01$ at $99\%$ confidence (\autoref{app:sizing}).
In all five runs, the observed reduction in tail optimality
gaps is larger than the improvement guaranteed by
$\CVaR_{0.95}(H'-H)$, showing that the bound is conservative.

\subsection{Model Capability and Search Architecture}
\label{sec:eval-methods}

We examine how model capability and search architecture
affect certification. Holding paired-CVaR feedback
fixed, we compare three models (GPT-4o-mini, Sonnet 5.5,
and Opus 5) across five harnesses: Engram~\cite{karimi2026engram}, Engram with adversarial training inputs
(\autoref{app:adversarial}),
Robusta's routed specialists~\cite{karimi2025robusta}, the evolutionary search method
ShinkaEvolve~\cite{shinkaevolve2025}, and Sequential Repair.
Sequential Repair starts with an Engram-generated heuristic
and uses successive sessions of coding-agents to identify
regions where it can improve upon the baseline and
propose targeted repairs (see~\autoref{app:workflow}).

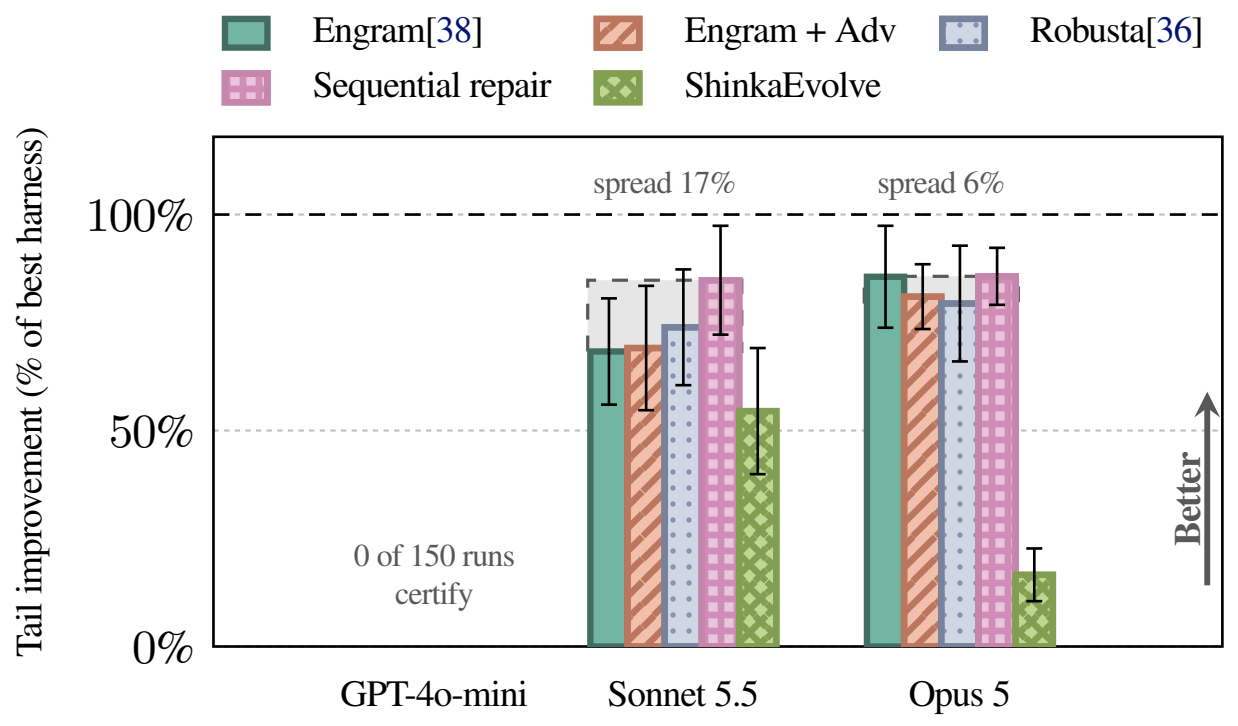
\begin{figure}[t]
\centering
\begin{tikzpicture}
\begin{axis}[barplot/axis, width=\linewidth, height=0.6\linewidth,
  set layers,
  bar width=7pt,
  cycle list={
    {bar shift=-2*\pgfplotbarwidth, draw=cGen!60!black, fill=cGen},
    {bar shift=-1*\pgfplotbarwidth, draw=cGenAdv!80!black, fill=cGenAdv!60,
     postaction={pattern={Lines[angle=45,distance=3pt]}, pattern color=cGenAdv!80!black}},
    {bar shift=0*\pgfplotbarwidth, draw=cRouted!80!black, fill=cRouted!45,
     postaction={pattern={Dots[distance=3pt]}, pattern color=cRouted!90!black}},
    {bar shift=+1*\pgfplotbarwidth, draw=cRepair!90!black, fill=cRepair!55,
     postaction={pattern={Hatch[angle=0,distance=3pt]}, pattern color=cRepair!95!black},
     postaction={pattern={Hatch[angle=90,distance=3pt]}, pattern color=cRepair!95!black}},
    {bar shift=+2*\pgfplotbarwidth, draw=cEvo!80!black, fill=cEvo!70,
     postaction={pattern={Hatch[angle=45,distance=3pt]}, pattern color=cEvo!80!black}}},
  xtick={0.1,1,2}, xticklabels={GPT-4o-mini, Sonnet 5.5, Opus 5},
  xmin=-0.3, xmax=2.55,
  xticklabel style={font=\scriptsize},
  ymin=0, ymax=118, ytick={0,50,100}, yticklabel={\pgfmathprintnumber{\tick}\%},
  ylabel={Tail improvement (\% of best harness)},
  ylabel style={align=center, font=\scriptsize},
  extra y ticks={100}, extra y tick labels={},
  extra y tick style={grid=major, grid style={black, dashed, line width=0.5pt}},
  error bars/y dir=both, error bars/y explicit,
  error bars/error bar style={line width=0.5pt, black},
  error bars/error mark options={rotate=90, mark size=1.5pt, line width=0.5pt},
  legend columns=3,
  legend style={at={(0.5,1.03)}, anchor=south, draw=none, fill=none,
                font=\scriptsize, column sep=6pt},
]
\legend{Engram\cite{karimi2026engram}, Engram + Adv, Robusta\cite{karimi2025robusta}, Sequential repair, ShinkaEvolve}

\begin{pgfonlayer}{axis background}
  \filldraw[fill=set2gray!15, draw=set2gray, dashed, line width=0.6pt]
    ([xshift=-18pt]axis cs:1,68.3) rectangle ([xshift=11pt]axis cs:1,84.8);
  \filldraw[fill=set2gray!15, draw=set2gray, dashed, line width=0.6pt]
    ([xshift=-18pt]axis cs:2,79.4) rectangle ([xshift=11pt]axis cs:2,85.7);
\end{pgfonlayer}

\node[font=\tiny, set2gray, anchor=south, inner sep=1pt]
  at ([xshift=-3.5pt]axis cs:1,102) {spread 17\%};
\node[font=\tiny, set2gray, anchor=south, inner sep=1pt]
  at ([xshift=-3.5pt]axis cs:2,102) {spread 6\%};

\addplot coordinates {(1,68.3) +- (0,12.3) (2,85.6) +- (0,11.8)};
\addplot coordinates {(1,69.1) +- (0,14.4) (2,81.0) +- (0, 7.5)};
\addplot coordinates {(1,73.9) +- (0,13.4) (2,79.4) +- (0,13.4)};
\addplot coordinates {(1,84.8) +- (0,12.6) (2,85.7) +- (0, 6.6)};
\addplot coordinates {(1,54.5) +- (0,14.6) (2,16.6) +- (0, 6.1)};

\node[font=\tiny, set2gray, align=center, anchor=south] at (axis cs:0.1,4)
  {0 of 150 runs\\certify};
\betterV{0.985}{0.12}{0.5}
\end{axis}
\end{tikzpicture}
\caption{\textbf{Coding-agent harnesses converge with stronger models.}
Tail improvement relative to the best harness for each
benchmark and model ($100\%$ matches the best).
Bars average across six benchmarks, with standard errors.
The spread among the four coding-agent harnesses decreases
from $17\%$ with Sonnet 5.5 to $6\%$ with Opus 5.
GPT-4o-mini certifies none of its runs.}
\label{fig:ratio}
\vspace{-10 pt}
\end{figure}

\autoref{fig:ratio} shows with GPT-4o-mini, none of the 150 runs
certifies, regardless of harness. With stronger models, the four
coding-agent harnesses perform increasingly alike: their tail improvement spread relative to the best harness across problems shrinks from 17\% with Sonnet 5.5 to 6\% with Opus 5. ShinkaEvolve, the evolutionary harness, trails in both tail
improvement and certification rate compared to coding-agent harnesses.

\begin{takeaway}
In our experiments, as models became more capable, the difference between simple and advanced harnesses became less pronounced. However, the choice of objective still had a significant impact.
\end{takeaway}

\section{Discussion and Limitations}
\label{sec:limits}
\label{sec:discussion}

\noindent\textbf{Training-time constraints do not guarantee SLO compliance.}
\name\ optimizes solution quality, while enforcing code-execution latency
constraints by rejecting candidates that exceed the allowance
on training workloads. However, on unseen $k$-median certification workloads,
many generated heuristics exceed latency allowance (\autoref{fig:latency}). This suggests that latency
should be included directly in the synthesis objective,
alongside tail performance, rather than enforced only
through training-time filtering.

\noindent\textbf{Certificates are relative to a baseline.}
Two heuristics certified on the same baseline aren’t always comparable and should be directly evaluated.
However, certificates can be chained: if $H'$ certifies against $H$, and $H''$ certifies against $H'$, then $H''$ also improves over $H$, assuming the same workload and tail level (\autoref{app:relative}).
This lets a certified heuristic serve as the baseline for future synthesis.

\noindent\textbf{Certification is distribution-specific.}
\name's guarantees apply only to the workload distribution
used for certification and may not hold under distribution
shifts. A new distribution requires re-certification and,
potentially, further synthesis. We leave
robustness under distribution shifts to future work.

\noindent\textbf{Certification requires regression bounds.}
Statistical certification requires an upper bound $M$ on
how much worse a generated heuristic can perform than
the baseline over the distribution.
We currently derive these bounds manually, which
can be hard and may yield overly conservative
certificates. We leave automating this to future work.

\section{Related Work}
\label{sec:related}

\noindent\textbf{LLM-based heuristic design.}
FunSearch, EoH, ReEvo, LLaMEA, AlphaEvolve, and ShinkaEvolve use executable feedback to search for programs and heuristics
\cite{romeraparedes2024mathematical,liu2024eoh,ye2024reevo,stein2025llamea,novikov2025alphaevolve,shinkaevolve2025}.
Agentic frameworks, including AIDE, AHD Agent, Glia, and Engram, use iterative experimentation and code refinement
\cite{jiang2025aide,lv2026ahdagent,glia,karimi2026engram}.
ADRS, SkyDiscover, AdaEvolve, and EvoX apply AI-driven search to systems problems
\cite{cheng2025barbarians,liu2026skydiscover,cemri2026adaevolve,liu2026evox}. HeurAgenix and EoH-S incorporate ideas from algorithm selection
\cite{yang2025heuragenix,liu2026eohs,rice1976algorithm,xu2008satzilla,hutter2011sequential}.
Rather than proposing a new search architecture, \name focuses on the synthesis objective and certifying tail improvements relative to a baseline.

\noindent\textbf{Heuristic analysis.}
MetaOpt searches for adversarial inputs that expose performance gaps between heuristics or against the optimum~\cite{namyar2024metaopt}.
XPlain explains heuristic failures, while MetaEase uses source-code analysis to discover adversarial inputs~\cite{xplain,karimi2026metaease}.
AIChilles identifies regressions in AI-generated programs and incorporates them into synthesis feedback~\cite{zhou2026aichilles}.
While these methods identify failures, \name proposes a scalable synthesis objective with guarantees that measures failure frequency and severity over a workload distribution and bounds changes in tail optimality gaps without computing OPT.

\noindent\textbf{Risk measures and robustness.}
CVaR and its coherence properties are well established
\cite{artzner1999coherent,rockafellar2000optimization,rockafellar2002conditional,acerbi2002coherence}.
TeaVaR uses CVaR to balance WAN throughput and availability~\cite{bogle2019teavar}, while Learn then Test provides finite-sample risk guarantees~\cite{angelopoulos2025learnthen}.
Learning-augmented algorithms study robustness to prediction errors
\cite{lykouris2021competitive,purohit2018improving,mitzenmacher2022algorithms,lechowicz2024online}, and risk-sensitive online algorithms also consider CVaR~\cite{christianson2024risk}. \name uses these concepts to provide systematic certification for distributional tail improvements in LLM-based heuristic design.

\section{Conclusion}

In this paper, we studied how to guide LLMs to design heuristics that improve tail performance without sacrificing average performance. We introduced \name, which uses the CVaR of paired candidate--baseline cost differences as a scalable synthesis objective. We proved that a negative paired CVaR guarantees improvements in both average performance and tail optimality gaps, without computing the optimal solution. We developed finite-sample confidence bounds to certify these improvements on unseen workloads and a preflight check to rule out unattainable certification targets before synthesis. Across three case studies, \name achieves the best mean--tail trade-off among the evaluated synthesis objectives. Across seven benchmarks, it generates heuristics certified at 99\% confidence to improve both average performance and tail optimality gaps on six. On these six benchmarks, its best heuristics also outperform robust human-designed alternatives in the tail, with comparable or better average gains.

\clearpage
\bibliographystyle{mlsys2025}


\clearpage
\appendix

\section{Theoretical Guarantees and Statistical Certification}
\label{app:math}

This appendix presents the guarantees underlying \name.
We first establish the relationship between paired-CVaR
and tail optimality gaps, then describe the statistical
certification procedure and its application to our benchmarks.

\subsection{Definitions and Assumptions}
\label{app:defs}
\label{app:assump}

Let $I\sim\mathcal{D}$ denote a workload drawn from the
target distribution. We express all problems as cost
minimization. We assume that $H$, $H'$, and $\OPT$ are
integrable and that both heuristics return feasible
solutions, so
\[
\OPT(I)\leq\min\{H(I),H'(I)\}
\]
for every workload $I$.

The two heuristics are evaluated on the same workload;
their costs need not be independent.

For an integrable random variable $X$ and $0<\beta<1$,
CVaR can be expressed equivalently as
\begin{equation}\label{eq:ru-var}
\begin{aligned}
\CVaR_\beta(X)
&=\min_{t\in\mathbb{R}}
\left\{
t+\frac{\mathbb{E}[(X-t)^+]}{1-\beta}
\right\}\\
&=\frac{1}{1-\beta}\int_\beta^1 \VaR_u(X)\,du.
\end{aligned}
\end{equation}
Here $z^+=\max\{z,0\}$ and $\VaR_u(X)$ denotes
the $u$-quantile of $X$
\cite{rockafellar2000optimization,rockafellar2002conditional}.
CVaR averages the upper $1-\beta$ fraction of the
distribution, accounting for probability mass at
the quantile boundary.

\subsection{The Optimum-Free Certificate}
\label{app:certificate-proof}
\label{app:subadd}

\begin{proposition}[Subadditivity]\label{prop:subadd}
For integrable $X,Y$,
$\CVaR_\beta(X+Y)\leq\CVaR_\beta(X)+\CVaR_\beta(Y)$.
\end{proposition}
\begin{proof}
For any $s,t\in\mathbb{R}$, $(X+Y-s-t)^+\leq(X-s)^++(Y-t)^+$, so the
variational expression in \autoref{eq:ru-var} for $X+Y$ at $s+t$ is at most
the sum of the expressions for $X$ at $s$ and $Y$ at $t$. Minimizing over
$s$ and $t$ separately gives the claim
\cite{artzner1999coherent,acerbi2002coherence}.
\end{proof}

\begin{theorem}[Optimum-free tail certificate]\label{thm:chain}
Under the assumptions of \autoref{app:assump}, for any $0<\beta<1$,
\[
\begin{aligned}
&\CVaR_\beta(H'-\OPT)
-\CVaR_\beta(H-\OPT)\\
&\qquad\leq\CVaR_\beta(H'-H).
\end{aligned}
\]
\end{theorem}
\begin{proof}
Write $H'-\OPT=(H-\OPT)+(H'-H)$, apply \autoref{prop:subadd}, and
subtract the finite quantity $\CVaR_\beta(H-\OPT)$.
\end{proof}

\begin{corollary}[Average improvement]\label{cor:mean-improvement}
$\mathbb{E}[H'-H]\leq\CVaR_\beta(H'-H)$ for $0<\beta<1$. A negative population
certificate therefore also implies $\mathbb{E}[H']<\mathbb{E}[H]$.
\end{corollary}
\begin{proof}
The quantile function ($\VaR$) is nondecreasing, so its average over the upper
$1-\beta$ mass in \autoref{eq:ru-var} is at least its average over
$[0,1]$, which is the mean.
\end{proof}

\subsection{Certification as a Partial Order}
\label{app:relative}

Fix a distribution $\mathcal{D}$ and tail level $\beta$.
We say that $B$ improves on $A$, written $B\prec A$, if
$\CVaR_\beta(B-A)<0$.

This relation is a strict partial order: a heuristic cannot
improve on itself, improvements are transitive, and two
heuristics cannot both improve on each other.
These properties follow from $\CVaR_\beta(0)=0$ and
subadditivity. In particular, if $C\prec B$ and $B\prec A$,
then
\[
\CVaR_\beta(C-A)
\leq \CVaR_\beta(C-B)+\CVaR_\beta(B-A)<0.
\]

\subsection{The Empirical $\CVaR$ Estimator}
\label{app:estimator}

In practice, we observe only a finite sample. For observations $x_{(1)}\leq\cdots\leq x_{(n)}$ and
$k=\lceil\beta n\rceil$,
\begin{equation}\label{eq:ru}
\widehat{\CVaR}_\beta(x)=
\frac{\sum_{i>k}x_{(i)}+(k-\beta n)\,x_{(k)}}{(1-\beta)n}.
\end{equation}

For a finite set of observations, \autoref{eq:ru} is \autoref{eq:ru-var} applied to the empirical measure, so
\autoref{thm:chain} holds exactly on any finite paired sample.

\subsection{Statistical Certification}
\label{app:statistical-procedure}
\label{app:uncertainty}
\label{app:sizing}
\label{app:statistical-procedure}

\paragraph{Certification sample size.}
After synthesis, we fix the selected candidate $H'$
and evaluate it alongside the baseline $H$ on an
independent certification sample $S$ of size $n$.
These workloads are not used during synthesis or
candidate selection.
Since $\CVaR_\beta$ measures the average of the worst
$1-\beta$ fraction of outcomes, we need sufficiently many
observations to estimate this tail.

We use the one-sided Dvoretzky--Kiefer--Wolfowitz (DKW) inequality
\cite{dvoretzky1956asymptotic,massart1990tight}
to guide sample sizing, which bounds the upper tail error by:
\[
\Pr\left(
\VaR_\beta \leq \widehat{\VaR}_{\beta+\eta}
\right)
\geq 1 - e^{-2n\eta^2}.
\]
For an upper confidence bound with probability $1-\delta$, setting the right-hand side to $1-\delta$ gives the DKW error margin $\varepsilon=\sqrt{\ln(1/\delta)/(2n)}$.
To achieve a desired quantile resolution $\varepsilon\leq\eta$, we therefore need:
\begin{equation}
\label{eq:sample-size}
n\geq\frac{\ln(1/\delta)}{2\eta^2}.
\end{equation}
Thus, higher confidence (smaller $\delta$) or finer
quantile resolution (smaller $\eta$) requires more
samples. We use $\delta=0.01$ throughout, so achieving $\varepsilon\leq0.01$ requires
$n\geq23{,}026$.

\paragraph{Finite-sample upper bound on CVaR.}
Even with many samples, a negative empirical paired CVaR
does not guarantee a negative population CVaR: some
rare, severe regressions may not appear in the sample.
We therefore use DKW to derive an upper confidence
bound on the true paired CVaR.

Let $\Delta=H'-H$, and suppose $\Delta(I)\leq M$
for every workload $I$ in the declared distribution.
With probability at least $1-\delta$, the one-sided
DKW inequality guarantees
\[
\VaR_u(\Delta)
\leq\widehat{\VaR}_{u+\varepsilon}(\Delta),
\qquad u\leq1-\varepsilon.
\]
This bounds the population quantiles using empirical
quantiles shifted by $\varepsilon$. For the highest
$\varepsilon$ fraction, which may contain outcomes
missing from the sample, we instead use the worst-case
regression bound $M$.

Since CVaR is the average of the upper-tail quantiles,
integrating these bounds gives, for
$\varepsilon\leq1-\breq$,
\begin{equation}
\label{eq:dkw-cvar}
\begin{aligned}
U_{\mathrm{DKW}}
&=\frac{1}{1-\breq}
\left(
\int_{\breq}^{1-\varepsilon}
\widehat{\VaR}_{u+\varepsilon}(\Delta)\,du
+\varepsilon M
\right)\\
&=\frac{
(1-\breq-\varepsilon)
\widehat{\CVaR}_{\breq+\varepsilon}(\Delta)
+\varepsilon M
}{1-\breq}.
\end{aligned}
\end{equation}
The first term measures the observed tail, while the
second accounts for potentially unseen regressions.
More samples reduce $\varepsilon$, while a tighter
$M$ reduces the penalty for unseen outcomes.

\paragraph{Certification guarantee.}
For a fixed candidate and an independent certification
sample $S$, the bound satisfies
\begin{equation}
\label{eq:dkw-guarantee}
\Pr_S\left[
\CVaR_{\breq}(\Delta)
\leq U_{\mathrm{DKW}}(S)
\right]\geq1-\delta.
\end{equation}
If $U_{\mathrm{DKW}}<0$, we certify improvements in
both mean performance and tail optimality gaps,
by \autoref{thm:chain} and
\autoref{cor:mean-improvement}.
For each fixed candidate, the probability of
incorrectly certifying a nonnegative population
paired CVaR is at most $\delta$.
We use $\delta=0.01$, giving 99\% confidence
per candidate. 

\paragraph{Bounding regressions ($M$)}
Certification requires an upper bound $M$ on
$\Delta(I)=H'(I)-H(I)$ for every workload in the
target distribution. We derive these bounds from
the generated programs and workload constraints,
rather than from observed regressions.

\emph{Offline Problems.}
For offline problems, a heuristic can compute a
candidate solution and compare it with the baseline
before returning it. If it keeps the candidate only
when it is no worse than the baseline, then
$\Delta(I)\leq 0$ and $M=0$.
This applies to all five VBP2D programs and three
programs each for TE MLU and $k$-median.

For the remaining two $k$-median programs, which
do not compare their solutions against the baseline,
we use a bound that applies to any feasible solution.
Each workload contains at most 70 sites in
$[0,1000]^2$, of which at least three are selected
as servers. Under the Manhattan metric, each
unselected site contributes at most 2000 to the
total service cost. Thus, the maximum possible cost,
and hence regression, is
$M=(70-3)\times2000=1.34\times10^5$.

The remaining two TE MLU programs (reps 1 and 4)
also compare against the baseline, but their MLU
calculations can differ slightly from the evaluation
harness due to floating-point rounding. They retain
the baseline routing unless their own calculation
indicates an improvement, subject to a tolerance
of $10^{-9}$. We bound the numerical disagreement
using at most $K=5.52\times10^6$ arithmetic operations,
unit roundoff $u=2^{-53}$, and a maximum ratio of
total to minimum link capacity of 534.6 (Cogentco).
The resulting disagreement is below $3.3\times10^{-7}$.
Since the harness reports MLU to six decimal places,
we conservatively use $M=10^{-6}$.

\emph{Online Problems.}
Unlike offline heuristics, online policies cannot
undo earlier decisions. We therefore bound the
largest cumulative loss each generated program
can incur relative to the baseline.

In Online Weighted Matching, each stream consists
of groups of arriving edges. Two programs can make
a worse decision only in the first group, after
which they follow the observed regime. Their maximum
regression is therefore bounded by that group's
weight, $M=4$. The other three programs can lose
weight in every group. Bounding their losses over
the longest possible streams gives
$M\in\{336,328,328\}$.

In GPU Batch Admission, each program's decisions
depend only on the request value level (1--8),
the remaining capacity, their weight, and the number of rejected
requests. This finite state space allows us to use
dynamic programming to find the maximum possible
loss against FCFS over all valid request streams.
The resulting bounds are
$M\in\{0.60,0.65,0.80,1.10,2.25\}\times10^6$.

For Online VBP2D, two out of five programs reverted to $H$, so $M=0$. The other three programs use
at most $n$ bins, where $n\leq4\OPT$ and
$\OPT\leq80$. Their regression is therefore
bounded by $n-\OPT\leq3\OPT\leq240$, giving
$M=240$.

Azure VM Packing uses a finite population instances (15 machine groups
over 96 hours). We evaluate every instance and
compute the population $\CVaR_\beta(\Delta)$ exactly,
so no statistical confidence bound is needed.

\paragraph{Certification Results.}
\autoref{tab:dkw} summarizes the five synthesis
runs per benchmark. For sampled workloads, $U$
denotes the 99\% DKW upper confidence bound on
paired CVaR. For Azure VM Packing, $U$ is the
exact population CVaR. A negative $U$ certifies
improvements in both average performance and
tail optimality gaps.

\begin{table}[t]
\centering
\small
\caption{\textbf{Certification at $\beta=0.95$.}
Range of $U$ across five synthesis runs, normalized by
the magnitude of the baseline's mean cost.
For sampled distributions, $U=U_{\mathrm{DKW}}$ with
$\delta=0.01$; for Azure VM Packing, $U$ is the
exact finite-population CVaR. Negative values
indicate certified improvements.}
\label{tab:dkw}
\resizebox{\linewidth}{!}{
\begin{tabular}{@{}lrr@{}}
\toprule
Benchmark & $U$ (\%) & Certified \\
\midrule
Metric $k$-median & $-0.22$ to $+371.68$ & 3/5 \\
Online VBP2D & $0.00$ to $+93.07$ & 0/5 \\
Online weighted matching & $-0.42$ to $+5.54$ & 2/5 \\
VBP2D & $-0.29$ to $-0.18$ & 5/5 \\
TE MLU & $-0.32$ to $0.00$ & 4/5 \\
GPU Batch Admission (Azure) & $-4.23$ to $-2.80$ & 5/5 \\
Azure VM Packing & $-5.17$ to $-4.85$ & 5/5 \\
\midrule
Total & & 24/35 \\
\bottomrule
\end{tabular}
}
\end{table}

\section{Certification Bounds}
\label{app:bounds}

\subsection{Candidate-Independent Floor and Ceiling}
\label{app:floor-ceiling}

Before searching for a better heuristic, we can determine
two limits: how much improvement is theoretically possible,
and when certification is impossible.

\paragraph{How much improvement is possible?}
Suppose we know a lower bound $\LB\leq\OPT$ on the
optimal cost. Since no feasible candidate can
perform better than $\OPT$, we obtain the following bound.

\begin{theorem}[Certificate floor]\label{thm:floor}
For any feasible candidate $H'$,
\begin{equation}\label{eq:certificate-floor}
\begin{split}
\CVaR_\beta(H'-H)
&\geq \CVaR_\beta(\OPT-H) \\
&\geq \CVaR_\beta(\LB-H) =: F_\beta.
\end{split}
\end{equation}
\end{theorem}

\begin{proof}
Since $H'\geq\OPT\geq\LB$, subtracting $H$
preserves these inequalities. The result follows
from the monotonicity of CVaR.
\end{proof}

The floor $F_\beta\leq0$ tells us how negative
the certificate could possibly become.
When $F_\beta<0$, we measure how much of this
potential improvement a candidate achieves:

\begin{equation}\label{eq:captured-fraction}
\frac{|\min\{0,\CVaR_\beta(H'-H)\}|}{|F_\beta|}
\in[0,1].
\end{equation}

A value closer to one means the candidate is closer
to the theoretical floor. However, this floor may
be unreachable, especially if $\LB$ is loose.

\paragraph{When is certification impossible?}
If the baseline already matches $\LB$ on many instances,
there may not be enough room for a negative tail
certificate.

\begin{corollary}[Certification ceiling]\label{cor:lbcap}
Let $\bceil=1-\Pr[H=\LB]$.
For any $\beta\geq\bceil$, no feasible candidate
can achieve a negative CVaR certificate.
\end{corollary}

\begin{proof}
Since $\LB-H\leq0$, and equals zero whenever
$H=\LB$, its worst $1-\beta$ fraction consists
entirely of zeros when
$1-\beta\leq\Pr[H=\LB]$.
Thus $F_\beta=0$. By \autoref{thm:floor},
every feasible candidate has
$\CVaR_\beta(H'-H)\geq0$.
\end{proof}
\subsection{Preflight with Sampling Confidence}
\label{app:preflight}

Draw $n$ i.i.d.\ preflight instances and let $K$ be the number on which
$H$ attains $\LB$. The one-sided Clopper--Pearson~\cite{clopper1934use} lower bound on
$p_0=\Pr[H=\LB]$ at confidence $1-\delta$ is
\[
p_{\mathrm{lo}}=
\begin{cases}
0,&K=0,\\
\mathrm{Beta}^{-1}(\delta;\,K,\,n-K+1),&K>0,
\end{cases}
\]
where $\mathrm{Beta}^{-1}$ is the beta quantile.
If $p_{\mathrm{lo}}>1-\breq$, the requested level is ruled out at that
confidence. The binomial coverage requires i.i.d.\ sampling and exact
equality $H=\LB$ under a valid lower bound; stratified or weighted
preflight samples need inference matched to that design.

\subsection{When Is Certification Impossible or Inconclusive?}
\label{sec:eval-preflight}

The Online VBP2D result raises a practical question: does a failed
certification indicate an unattainable target or simply an unsuccessful
synthesis attempt? Preflight can establish some impossibility cases
without launching synthesis, but passing preflight does not prove
that a certifiable candidate exists.

\textbf{An unattainable paired-CVaR target (Makespan).}
For Makespan, the LPT baseline matches a valid
lower bound on 35.49\% of sampled workloads,
giving an estimated certification ceiling
$\widehat{\bceil}=0.6451$.
At $\breq=0.95$, certification can be ruled out
when the lower confidence bound on this fraction
exceeds 5\%.
The preflight test evaluates this condition before
synthesis, without computing $\OPT$.
The human MULTIFIT control can nonetheless improve average
performance. Thus preflight rules out this \emph{particular certificate},
not all possible performance improvements.

\textbf{A target not ruled out, but not achieved (Online VBP2D).}
Here the estimated ceiling is $0.9997$, so the preflight condition
does not rule out a certificate at $\breq=0.95$. Nevertheless,
none of the tested synthesis methods obtains one. This is an
inconclusive search outcome, not proof of impossibility. More broadly,
the ceiling is a sufficient obstruction to a negative paired-CVaR
certificate, not a necessary condition for tail optimality-gap
improvement. 

\subsection{Validation Against the True Optimum}
\label{app:kmedian-truth}

We validate the paired CVaR on small-scale Metric $k$-Median instances for Engram's heuristics.
We solve 28,000 certification instances to optimality
using Gurobi~\cite{gurobi2026manual}, and compute LP
lower bounds using GLOP~\cite{google2026glop}.
The baseline, GreedyAdd, matches the lower bound on
0.767\% of instances, giving a certification ceiling of $\bceil = 0.9923$.

\autoref{tab:kmedian-truth} compares the paired CVaR
with the true change in tail optimality gap across five
Opus runs. The theoretical bound (\autoref{thm:chain})
holds in every run, including at $\beta=0.90$ and $0.99$
(not shown). The paired CVaR is conservative because
the worst paired differences and the worst optimality
gaps need not occur on the same instances.

\begin{table}[t]
\centering
\caption{Small-scale Metric $k$-Median validation
at $\beta=0.95$ using Engram (five Opus runs).
The baseline tail optimality gap is 2,173.
}
\label{tab:kmedian-truth}
\footnotesize
\setlength{\tabcolsep}{4pt}
\begin{tabular}{crrr}
\toprule
Run & Paired CVaR & True change & $H'=\OPT$ \\
\midrule
1 & $-6.3$  & $-1\,685.4$ & 79.1\% \\
2 & $-27.0$ & $-2\,117.3$ & 94.9\% \\
3 & $-12.7$ & $-1\,734.4$ & 69.6\% \\
4 & $-27.4$ & $-2\,142.8$ & 95.5\% \\
5 & $-27.7$ & $-2\,156.9$ & 97.6\% \\
\bottomrule
\end{tabular}
\end{table}

\section{Benchmarks and Experimental Methodology}
\label{app:workflow}
\label{app:eval_detail}

\autoref{tab:distributions} summarizes the workload distributions
used in our synthetic benchmarks. Certification uses independently sampled
instances not seen during synthesis.

\begin{table}[t]
\centering
\caption{\textbf{Baselines and human-designed comparisons.}
Makespan is ruled out in preflight.}
\label{tab:eval-benchmarks}
\footnotesize
\setlength{\tabcolsep}{3pt}
\renewcommand{\arraystretch}{1.15}
\begin{tabularx}{\linewidth}{@{}
    >{\raggedright\arraybackslash}p{0.33\linewidth}
    >{\raggedright\arraybackslash}p{0.23\linewidth}
    >{\raggedright\arraybackslash}X@{}}
\toprule
\textbf{Problem} & \textbf{Baseline} & \textbf{Human control} \\
\midrule
GPU Batch Admission 
    & FCFS & Zhou et al.~\cite{zhou2008budget} \\
\midrule
Azure VM packing
    & Tight packing & $L_2$ best-fit~\cite{panigrahy2011heuristics} \\
\midrule
Server Placement (Metric $k$-median)
    & GreedyAdd & 2-swap local search~\cite{arya2004local} \\

\addlinespace[4pt]
\cmidrule(lr){1-3}
\addlinespace[2pt]
Online weighted matching
    & Accept-if-free & Probe-based reservation \\
\midrule
VBP2D
    & First Fit & LP rounding~\cite{kulik2023improved} \\
\midrule
Online VBP2D
    & Online First Fit
    & Weighted $L_2$ fit~\cite{panigrahy2011heuristics} \\
\midrule
TE MLU
    & Shortest path & Greedy load balancing \\

\addlinespace[4pt]
\cmidrule(lr){1-3}
\addlinespace[2pt]
Makespan
    & LPT~\cite{graham1969bounds}
    & MultiFit + LPT guard~\cite{coffman1978application} \\
\bottomrule
\end{tabularx}
\end{table}

\subsection{Cost of the Offline Optimum}
\label{app:opt}

\name needs no optimum, but a reader may ask whether one could simply
compute it on large scale. \autoref{tab:opt-complexity} lists each benchmark's
offline problem, its hardness, an exact algorithm, the lower-bound we used in the preflight and its time complexity.
Computing the optimum does not scale to realistic instances.
For VBP2D, exhaustively evaluating all $10^{10}$
five-item sequences took 1.8 minutes on 16 cores,
but evaluating all $10^{20}$ ten-item sequences
is projected to take $10^5$ years!

\subsection{Synthesis Architectures and Adversarial Inputs}
\label{app:adversarial}
\label{app:architecture}

\begin{figure*}[tp]
\centering
\resizebox{0.94\linewidth}{!}{\input{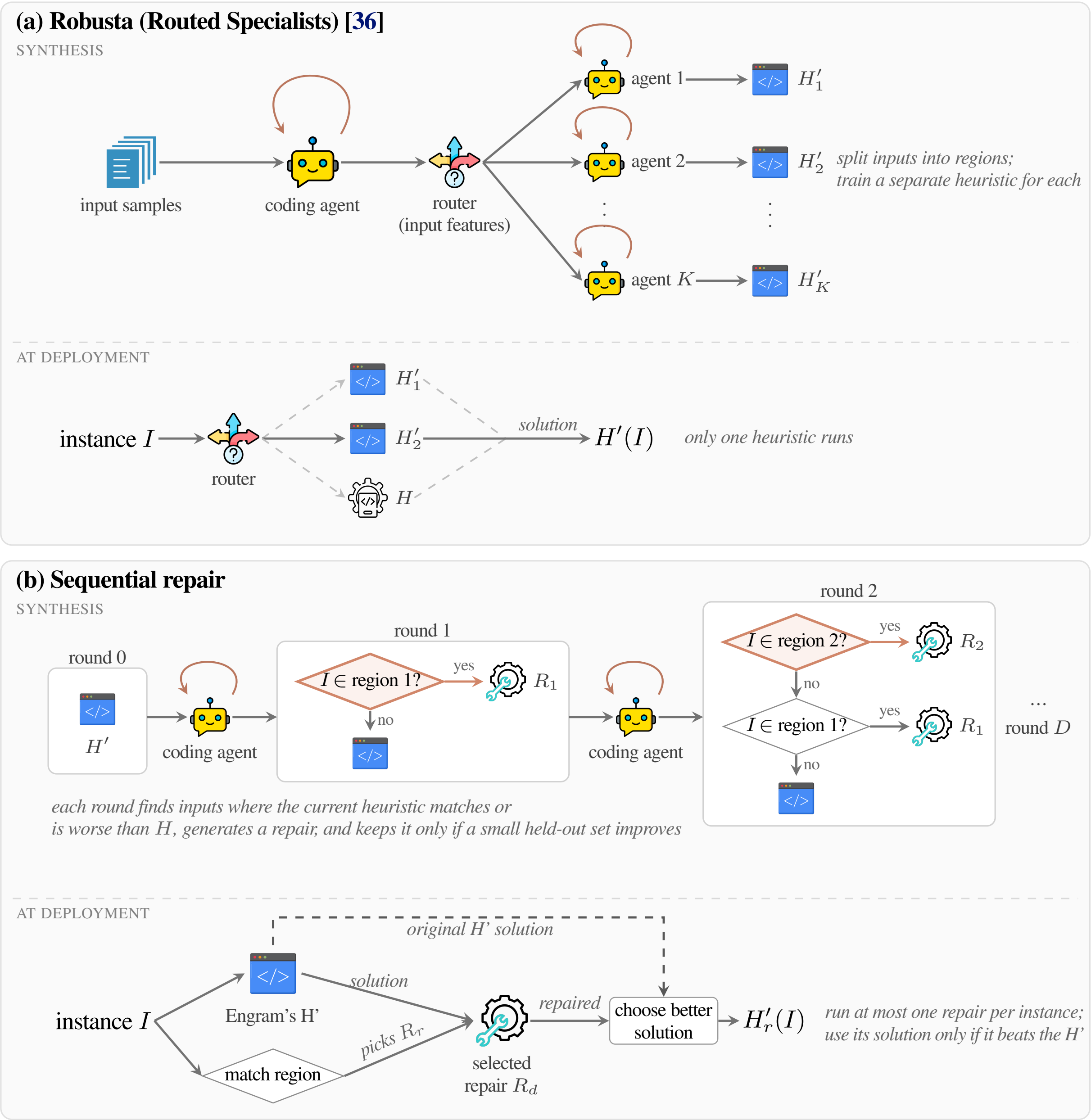}}
\caption{
\textbf{Candidate architectures.}
Each robot represents Engram~\cite{karimi2026engram},
where successive coding-agent sessions build on the
programs produced by earlier sessions.
(a) \textbf{Routed specialists} divide the input space
into regions and independently search for a heuristic
($H'_1,\ldots,H'_K$) for each region.
At deployment, the router selects one heuristic per instance.
(b) \textbf{Sequential repair} starts with an Engram-generated
heuristic and iteratively adds targeted repairs for regions
where the candidate still matches the baseline despite
room for improvement indicated by the lower bound.
At deployment, at most one repair runs per instance,
and its solution is used only if it improves upon
the original.
}
\label{fig:arch}
\end{figure*}

\autoref{fig:arch} shows the candidate architectures
used in our experiments, showing how heuristics are
generated and how they operate at deployment.

We generate the adversarial examples from two sources. First we leverage the family of hard adversarial samples introduced by prior work~\cite{namyar2024metaopt,karimi2026metaease}. Second, we modify AIChilles~\cite{zhou2026aichilles} and use an LLM to generate samples that provide large performance gaps against either
the exact optimum or a better feasible solution. Since the optimum is expensive, we cannot generate many such samples.

\clearpage

\begin{table*}[t]
\centering
\caption{Workload distributions used for synthetic benchmarks.}
\label{tab:distributions}
\footnotesize
\setlength{\tabcolsep}{5pt}
\begin{tabular}{@{}L{3cm}L{12cm}@{}}
\toprule
Benchmark & Workload distribution \\
\midrule

Online Weighted Matching
& 8--80 groups of weighted paths, with uniformly random arrival orders
and weights. \\
\addlinespace

VBP2D
& 16--400 items with randomly generated two-dimensional
resource demands and unit-capacity bins. \\
\addlinespace

Online VBP2D
& 8--80 perfectly packed bins, each split into 2--4 items
whose two-dimensional demands sum to one bin's capacity.
The resulting 16--320 items arrive in random order. \\
\addlinespace

TE MLU
& Five real network topologies with varying uniformly random traffic
patterns, load levels, and link failures. \\
\addlinespace

Makespan
& 2--26 machines with uniformly randomly generated job processing
times, plus constructed difficult instances. \\
\bottomrule
\end{tabular}
\end{table*}

\begin{table*}[t]
\centering
\caption{Offline optimum and preflight lower bound per benchmark. Every
bound is a relaxation of the offline optimum. $n$: items, jobs, clients, or vertices; $d$: demands;
$C$: knapsack capacity; $O^*$ hides polynomial factors.}
\label{tab:bounds}
\label{tab:opt-complexity}
\footnotesize
\setlength{\tabcolsep}{4pt}
\begin{tabular}{@{}L{2.1cm}L{5.2cm}L{6.6cm}l@{}}
\toprule
Benchmark & Offline optimum & Lower bound $\LB(I)$ & LB Time Complexity \\
\midrule
Server Placement (Metric $k$-median) &
  $k$ median; NP-hard~\cite{kariv1979pmedians,megiddo1984geometric};
  all center sets, $O(n^{k+1})$ &
  LP relaxation of the $k$-median program; integral, hence exact, on 92\%
  of instances &
  one LP \\
\addlinespace
GPU Batch Admission (Online Knapsack) &
  0/1 knapsack; weakly NP-hard~\cite{karp1972reducibility};
  DP, $O(nC)$~\cite{martello1990knapsack} &
  negated fractional-knapsack bound on captured value, filling in
  decreasing density order~\cite{dantzig1957discrete}
  &
  $O(n\log n)$ \\
\addlinespace
Online Weighted Matching &
  max-weight matching; polynomial,
  $O(n|E|+n^2\log n)$~\cite{edmonds1965paths,gabow1990matching}, but one
  solve per instance &
  $\tfrac12\sum_v\max_{e\ni v}w_e$ (negated for minimization): each vertex
  is matched at most once and each matched edge is counted at both ends;
  exact on the disjoint gadget grammar &
  $O(|E|)$ \\
\addlinespace
VBP2D, Online VBP2D &
  2D vector bin packing; strongly
  NP-hard~\cite{gareyjohnson1979computers,woeginger1997vector};
  subset DP, $O^*(3^n)$ &
  volume: $\max_d\lceil\sum_i s_{i,d}/C\rceil$, the bins the fullest
  dimension alone requires~\cite{martello1990knapsack}; exact on the
  constructed Online VBP2D groups &
  $O(n)$ \\
\addlinespace
TE MLU &
  single-path routing; NP-hard~\cite{kleinberg1996disjoint};
  all path choices, $O(4^d d)$ &
  fluid relaxation: split every demand freely over its path menu and
  minimize maximum link utilization~\cite{fortz2000internet} &
  one LP \\
\addlinespace
Makespan &
  $P\|C_{\max}$; strongly NP-hard~\cite{gareyjohnson1979computers};
  subset DP, $O^*(3^n)$ &
  $\max(\lceil\sum_j p_j/m\rceil,\ \max_j p_j)$: average load and longest
  job &
  $O(n)$ \\
\bottomrule
\end{tabular}
\end{table*}

\clearpage

\clearpage

\section{Additional Results}
\label{app:additional-results}

\subsection{Additional Results for Azure GPU Batch Admission}
\label{app:azure-motivation}

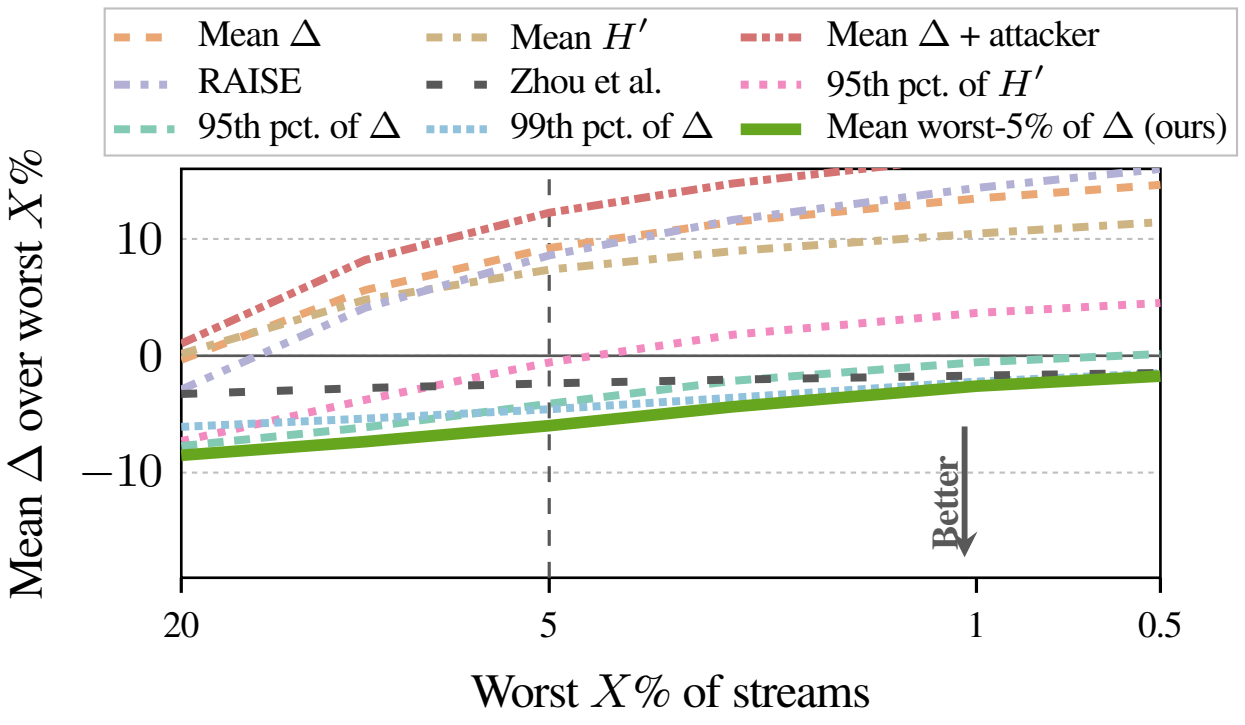
\begin{figure}[ht]
    \centering
\begin{tikzpicture}
\begin{axis}[paperaxis, width=0.47\textwidth, height=0.25\textwidth,
  every axis plot/.append style={line width=1.3pt, no markers},
  xmin=0.6990, xmax=2.3010, xtick={0.6990,1.3010,2.0000,2.3010},
  xticklabels={20,5,1,0.5}, xticklabel style={font=\scriptsize},
  restrict x to domain=0.6990:2.3010,
  xlabel={Worst $X$\% of streams}, ylabel={Mean $\Delta$ over worst $X$\%},
  ymin=-19, ymax=16,
  legend columns=3, legend cell align=left,
  legend style={font=\scriptsize, at={(0.5,1.03)}, anchor=south, draw=black!25,
    inner sep=1pt, nodes={inner sep=0.8pt}, /tikz/column sep=3pt},
  legend image code/.code={\draw[#1] (0cm,0cm) -- (0.4cm,0cm);}]
\draw[set2gray, line width=0.5pt] (axis cs:0.6990,0) -- (axis cs:2.3010,0);
\draw[set2gray, dashed, line width=0.6pt] (axis cs:1.3010,-19) -- (axis cs:1.3010,16);
\addlegendimage{objMean}  \addlegendentry{Mean $\Delta$}
\addlegendimage{objMeanH} \addlegendentry{Mean $H'$}
\addlegendimage{objAttack} \addlegendentry{Mean $\Delta$ + attacker}
\addlegendimage{objRaise} \addlegendentry{RAISE}
\addlegendimage{objZhou}  \addlegendentry{Zhou et al.}
\addlegendimage{objPctH}  \addlegendentry{95th pct.\ of $H'$}
\addlegendimage{objPct}   \addlegendentry{95th pct.\ of $\Delta$}
\addlegendimage{objPtail} \addlegendentry{99th pct.\ of $\Delta$}
\addlegendimage{objOurs}  \addlegendentry{Mean worst-5\% of $\Delta$ (ours)}
\addplot[objMean,  forget plot] table[x=x, y=cert_M] {figs/data/azm_beta_mean.dat};
\addplot[objMeanH, forget plot] table[x=x, y=cert_M] {figs/data/azm_beta_avg_pprime.dat};
\addplot[objPctH,  forget plot] table[x=x, y=cert_M] {figs/data/azm_beta_p95_pprime.dat};
\addplot[objPct,   forget plot] table[x=x, y=cert_M] {figs/data/azm_beta_p95_delta.dat};
\addplot[objPtail, forget plot] table[x=x, y=cert_M] {figs/data/azm_beta_p99_delta.dat};
\addplot[objAttack, forget plot] table[x=x, y=cert_M] {figs/data/azm_beta_mean_attackgap.dat};
\addplot[objRaise, forget plot] table[x=x, y=cert_M] {figs/data/azm_beta_raise.dat};
\addplot[objZhou,  forget plot] table[x=x, y=cert_M] {figs/data/azm_beta_zhou.dat};
\addplot[objOurs,  forget plot] table[x=x, y=cert_M] {figs/data/azm_beta_cvar.dat}; 
\betterV{0.8}{0.37}{0.05}
\end{axis}
\end{tikzpicture}
    \caption{\textbf{Tail improvements extend beyond the level used for
    synthesis.} Mean $\Delta$ over the worst $X\%$ of streams for each
    objective (lower is better). The dashed vertical line is $X=5\%$,
    the fraction optimized by paired CVaR. Its measured advantage persists
    across the plotted tail fractions, from 20\% to 0.5\%; these
    observations are not a guarantee for every tail level. }
    \label{fig:motivation-tail-sweep}
\end{figure}

\autoref{fig:motivation-tail-sweep} examines whether these
improvements extend beyond the 5\% tail used during
synthesis. Paired-CVaR maintains favorable measured
tail performance across worst-case fractions ranging
from 20\% to 0.5\%, including more extreme tails
than those explicitly optimized. While these results
suggest that the benefits are not limited to the
chosen tail level, they do not establish guarantees
at other tail levels.

\subsection{Training-Time Constraint Enforcement Does Not Guarantee SLO Compliance}
\label{sec:eval-latency}
\label{app:latency}

During synthesis, we enforce latency constraints by rejecting
candidates that exceed the latency allowance on training instances.
However, satisfying these constraints during training does not
guarantee compliance on larger, unseen workloads.

\begin{figure}[ht]
\centering
\begin{tikzpicture}
\begin{axis}[
  paperaxis,
  width=\linewidth,
  height=0.5\linewidth,
  xmin=3.6, xmax=7.6,
  ymin=-31, ymax=1.5,
  xtick={4,4.5,5,5.5,6,6.5,7,7.5},
  ytick={0,-5,-10,-15,-20,-25,-30},
  xlabel={Candidate p95 latency in isolation (ms)},
  ylabel={$\mathrm{CVaR}_{0.95}(\Delta)$ (service cost)},
  tick label style={font=\small},
  label style={font=\small},
  xlabel style={yshift=2pt},
  extra x ticks={5},
  extra x tick labels={},
  extra x tick style={
    grid=major,
    grid style={softcoral, solid, line width=0.8pt}
  },
  extra y ticks={0}, extra y tick labels={},
  extra y tick style={grid=major, grid style={black, solid, line width=0.5pt}},
  legend style={
    at={(0.5,1.03)}, anchor=south,
    font=\footnotesize,
    draw=black, fill=white,
    /tikz/every even column/.append style={column sep=6pt},
  },
  legend columns=2,
  legend cell align=left,
]
\node[font=\footnotesize, softcoral, rotate=90, anchor=south]
  at (axis cs:5,-10) {5 ms cutoff};

\addplot[only marks, mark=*, mark size=2.4pt, color=cGen, mark options={fill=cGen}] coordinates {
  (6.267,-19.13) (6.107,-28.29) (6.769,-19.09) (7.341,-27.98) (6.023,-28.16)
};
\addlegendentry{Engram~\cite{karimi2026engram}}

\addplot[only marks, mark=square*, mark size=2.2pt, color=cGenAdv, mark options={fill=cGenAdv}] coordinates {
  (4.846,-27.84) (5.612,-28.15) (6.127,-22.68) (6.527,-27.85) (4.972,-27.62)
};
\addlegendentry{Engram + Adversarial Samples}

\addplot[only marks, mark=triangle*, mark size=2.8pt, color=cRouted, mark options={fill=cRouted}] coordinates {
  (6.541,-2.45) (3.865,0.00) (6.256,-0.08) (5.460,-9.37) (5.650,-10.75)
};
\addlegendentry{Robusta~\cite{karimi2025robusta}}

\addplot[only marks, mark=diamond*, mark size=2.8pt, color=cRepair, mark options={fill=cRepair}] coordinates {
  (4.935,-27.86) (5.593,-28.16) (6.094,-23.19) (6.460,-27.85) (5.591,-27.57)
};
\addlegendentry{Sequential repair}

\betterD{0.24}{0.54}{0.06}{0.32}

\end{axis}
\end{tikzpicture}
\caption{\textbf{Performance certification does not imply constraint compliance.}
Each point shows the certificate and isolated p95 latency of a generated
metric $k$-median candidate.}
\label{fig:latency}
\end{figure}
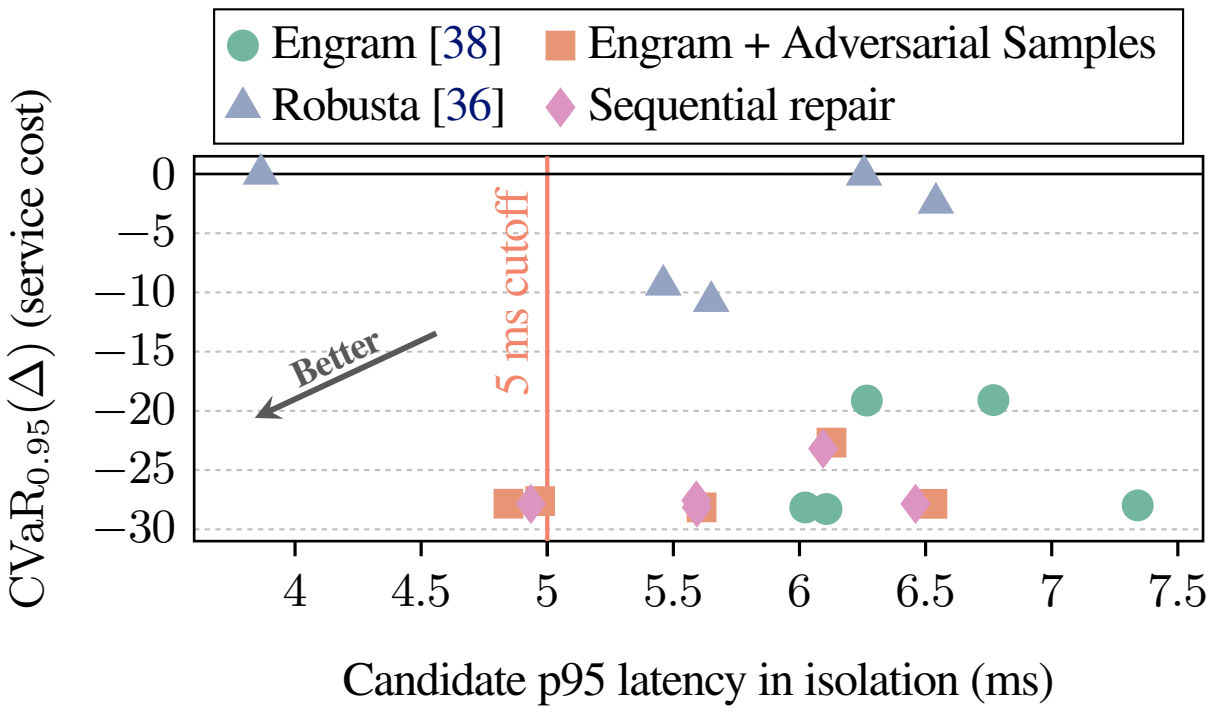

\autoref{fig:latency} shows this limitation for metric
$k$-median, where the latency allowance is
$\max(5\,\mathrm{ms}, 2.5\times\text{median baseline time})$.
Despite training-time filtering, most generated candidates exceed
the 5\,ms allowance in isolated evaluation, including candidates
that achieve substantial improvements in tail performance.

These results suggest that strong performance certificates do not
guarantee SLO compliance, motivating the incorporation of
constraints directly into the synthesis objective rather than relying solely
on training-time rejection of constraint violations.

\newsavebox{\arxiviconkeep}
\savebox{\arxiviconkeep}{%
  \includegraphics[height=1mm]{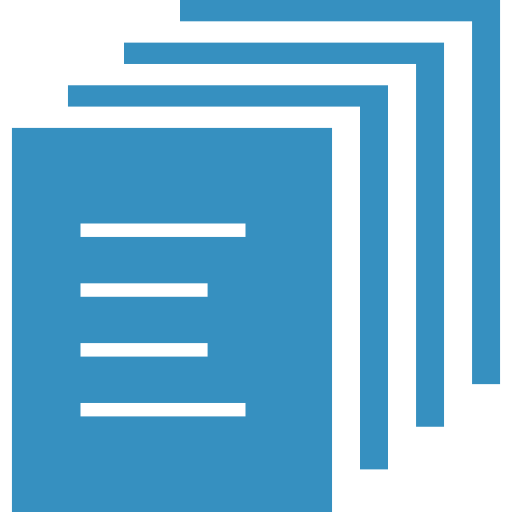}%
  \includegraphics[height=1mm]{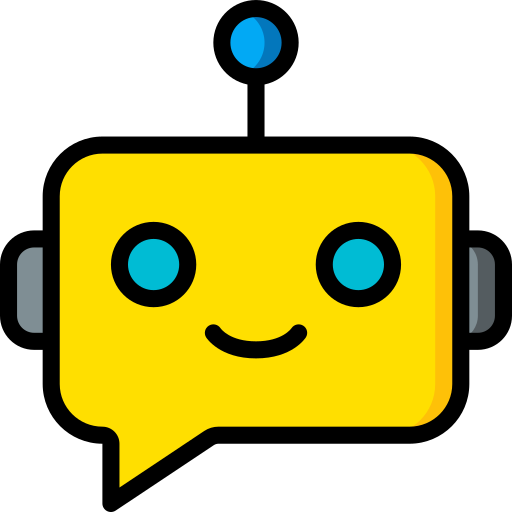}%
  \includegraphics[height=1mm]{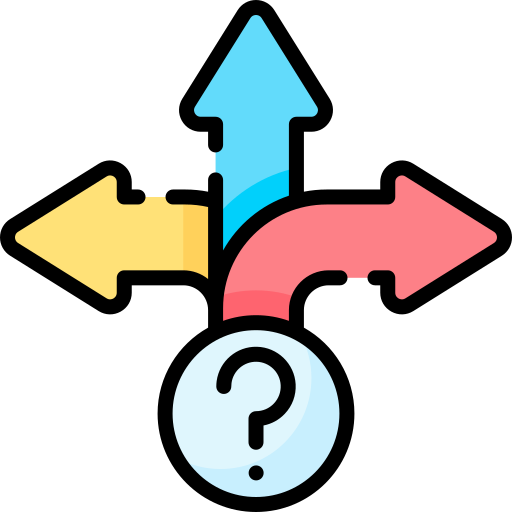}%
  \includegraphics[height=1mm]{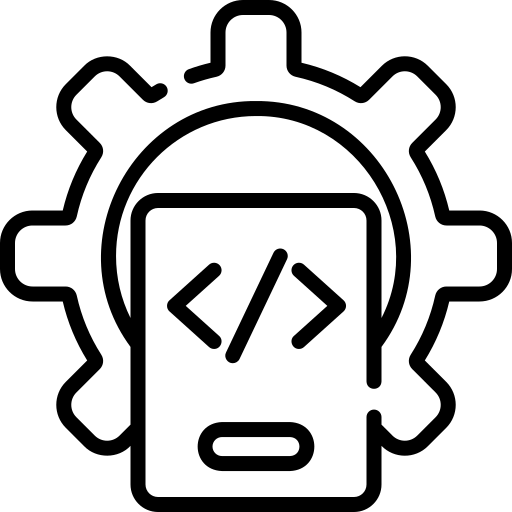}%
}

\end{document}